\pdfoutput=1
\documentclass[11pt]{article}

\usepackage[final]{acl}

\usepackage{times}
\usepackage{latexsym}

\usepackage[T1]{fontenc}
\usepackage[utf8]{inputenc}

\usepackage{microtype}

\usepackage{inconsolata}

\usepackage{graphicx}

\usepackage{microtype}
\usepackage{todonotes}
\usepackage{multirow}
\usepackage{setspace}
\usepackage{booktabs}
\usepackage{tabularray}
\usepackage{graphicx}
\usepackage{adjustbox}
\usepackage{amssymb}
\usepackage{CJKutf8}
\usepackage{xurl}
\usepackage[english,greek]{babel}

\newcommand{\jose}[1]{\textcolor{black}{#1}}
\newcommand{\dimos}[1]{\textcolor{black}{#1}}

\newcommand\DATASET{X-Topic}

\usepackage{caption}
\title{Multilingual Topic Classification in X: Dataset and Analysis}

\author{Dimosthenis Antypas$^1$, Asahi Ushio$^2$\thanks{Work done while at Cardiff NLP}, Francesco Barbieri$^3$, Jose Camacho-Collados$^1$
\\ $^1$Cardiff NLP, Cardiff University, United Kingdom $^2$Amazon, Tokyo, Japan \\ $^3$Snap Inc., Santa Monica, CA, USA \vspace{0.1cm}\\
$^1$\texttt{\{AntypasD,CamachoColladosJ\}@cardiff.ac.uk} 
$^2$\texttt{asahiu@amazon.com}
}

\begin{document}
\selectlanguage{english}
%\begin{spacing}{0.99}

\maketitle
\begin{abstract}
In the dynamic realm of social media, diverse topics are discussed daily, transcending linguistic boundaries. However, the complexities of understanding and categorising this content across various languages remain an important challenge with traditional techniques like topic modelling often struggling to accommodate this multilingual diversity.  In this paper, we introduce {\DATASET}, a multilingual dataset featuring content in four distinct languages (English, Spanish, Japanese, and Greek), crafted for the purpose of tweet topic classification. Our dataset includes a wide range of topics, tailored for social media content, making it a valuable resource for scientists and professionals working on cross-linguistic analysis, the development of robust multilingual models, and computational scientists studying online dialogue. Finally, we leverage {\DATASET} to perform a comprehensive cross-linguistic and multilingual analysis, and compare the capabilities of current general- and domain-specific language models.
\end{abstract}

\selectlanguage{english}
\section{Introduction}
Social platforms such as X (\textit{Twitter}), Snapchat and Instagram provide an environment for content creation and information sharing among people and organisations. In particular, people use these platforms to express their sentiments, share their opinions on multiple topics, and discuss and influence each other \cite{barbieri-etal-2014-modelling,hu2021revealing,ansari2020analysis}. In this scenario, these platforms are rich sources for informal short text, as they include content about recent events, shared by a heterogeneous group of users. The vast amount of content shared on social media, however, make it impossible to analyse and digest it without automatic tools.

Unsupervised approaches such as Latent Dirichlet Allocation (LDA) \cite{lda} and topic modelling variations \cite{steyvers2007probabilistic}, or more recently, BERTopic \cite{bertopic}, are common approaches to deal with this issue.
However, these methods are usually built as an ad-hoc analysis, with the derived topics not being easily interpretable or comparable among different analyses. On the other hand, when looking at supervising approaches, existing resources mainly focus on the news articles domain, e.g., BBC News \cite{greene2006practical}, Reuter \cite{lewis2004rcv1}, 20News \cite{lang1995newsweeder}, and WMT News Crawl \cite{lazaridou2021mind} with few exceptions like scientific (arXiv) \cite{lazaridou2021mind} and medical (Ohsumed) \cite{hersh1994ohsumed} domains.

Our paper focuses on expanding the resources available for multilingual tweet classification. %task, and specifically in a multi-label and multilingual setting. 
We leverage an initial topic taxonomy of 19 topics, first proposed in \newcite{antypas-etal-2022-twitter}, and introduce the new {\DATASET} dataset that includes tweets from four different languages: English, Spanish, Japanese and Greek. Our dataset is focused on \textit{X} data and aims to address the lack of labelled multilingual social media data, as well as to encourage the creation of new methods for multilingual topic classification. 

By leveraging {\DATASET} as a benchmark, we explore multiple model architectures and sizes for multilingual tweet \jose{topic} classification: %while establishing multiple experimental settings all based on multi-label classification, of different nature and difficulty:
(1) zero-shot, \jose{(2) few-shot,} (3) monolingual, (4) cross-lingual and (5) multilingual. Our analysis \jose{highlights the challenging nature of the task and} reveals interesting patterns in relation to the use of LLMs and supervised approaches for the topic classification task in social media, especially in relation to the type of data considered for training. 

\jose{The {\DATASET} dataset, as well as the topic classification models built upon it, are made openly available. {\DATASET} is available at \url{https://huggingface.co/datasets/cardiffnlp/tweet_topic_multilingual}. Table \ref{tab:examples} shows some sample instances of {\DATASET} for each language. Finally, the best multilingual models of \textit{base} and \textit{large} sizes are available at \url{https://huggingface.co/cardiffnlp/twitter-xlm-roberta-base-topic-multilingual} and \url{https://huggingface.co/cardiffnlp/twitter-xlm-roberta-large-topic-multilingual}, respectively.}

%CR 
%Overall, {\DATASET} can be a useful tool for researchers and industry experts to deal with challenges inherent in real-world applications in social media, especially given the relevance of topic analysis for extracting information and understanding online behaviour. {\DATASET} is especially suited to test multilingual solutions, specifically when dealing with languages from different families (e.g. English - Japanese) and also low resourced languages (Greek); as well as to consider new approaches that can fully utilise the latest advancements of large language models in low-resource settings. These challenges are built upon the existing problem of multi-label topic classification in social media where information and news are heavily diffused.

\begin{table*}[!ht]
\centering
\scalebox{0.8}{
\begin{tabular}{l|l}
\multicolumn{1}{c|}{\textbf{Tweet}} & \multicolumn{1}{c}{\textbf{Topics}} \\
\begin{tabular}[c]{@{}l@{}}\textbf{en}: I don’t think I really want to go to  Coachella  unless Taylor Swift is headlining\end{tabular} & Celebrity \& Pop Culture, Music \\ \hline
\begin{tabular}[c]{@{}l@{}}\textbf{es}: quiero una date en un museo\\ \\ \textbf{translation}: I want a date in a museum\end{tabular} & \textit{\begin{tabular}[c]{@{}l@{}}Relationships, Arts \& Culture,\\ Diaries \& Daily Life\end{tabular}} \\ \hline
\begin{CJK*}{UTF8}{min}\begin{tabular}[c]{@{}l@{}}\textbf{ja}: 久々になーーんもしないでいい日が二日もあるのでゆっくり富平井絆果と\\向き合うよ\\ \\ \textbf{translation}: It's been a long time since I've had two  days where I don't have to do anything, \\ so I'm going to take my time and face Kizuna Fuhirai.\end{tabular}\end{CJK*} & \textit{Diaries \& Daily Life, Gaming} \\ \hline
\begin{tabular}[c]{@{}l@{}}\textbf{gr}:\selectlanguage{greek} Μπα σε καλό σου μωρή Ανθουλα μας κοψοχολιασες πάλι \#σασμός\\ \\ \textbf{translation}: Oh my goodness, Anthula, you've cracked us up again  \#sasmos\end{tabular} & Film, TV \& Video
\end{tabular}
}
\caption{Example of tweets present in each language subset of \DATASET.}\label{tab:examples}
\end{table*}

\section{Related Work}

The task of classifying topics in social media content has garnered significant attention from the research community in recent years \cite{schlichtkrull2023intended,10.1145/3161603,chua2016linguistic}. Social media platforms like \textit{X} have become hubs for the exchange of information, opinions, and sentiments, making the development of effective classification methods imperative. 

\paragraph{Unsupervised Approaches.}
Due to the lack of labelled data and the dynamic nature of social media platforms, unsupervised methods have been widely used for topic modelling and classification on the content shared. Several variations of LDA have been introduced that try to address the challenges that arise when working with the often messy and unstructured world of social media. Such solutions, \cite{zhao2011comparing,rosen2004author,steinskog-etal-2017-twitter} often try to  combine author information with the text shared.
Other approaches use unsupervised clustering algorithms, such as k-means or hierarchical clustering, to group similar social media content based on their topic similarity \cite{10.1007/978-3-319-67256-4_30}. These methods are particularly useful when the underlying topics are not predefined and need to be inferred from the data. However, a drawback of these unsupervised approaches is that the derived topics may not always be easily interpretable \jose{or comparable across corpora}.

%\noindent{\textbf{Supervised Approaches.}}
%Supervised methods for topic classification in social media content involve training machine learning models on labelled data. While supervised approaches have demonstrated robust performance \cite{huang2013sentiment}, there is a notable scarcity of labelled data for social media content, particularly in languages other than English \cite{selvaperumal2014short}; while a lot of the available datasets offer a limited taxonomy of topics \cite{vadivukarassi2019comparison}. Finally, there is research that utilises semi-supervised approaches \cite{setiawan2016feature} or weakly labelled datasets \cite{yuan2018incorporating} to address the scarcity problem.

\paragraph{Multilingual resources in social media.}
Supervised methods for topic classification in social media content involve training machine learning models on labelled data. While supervised approaches have demonstrated robust performance \jose{on social media tasks} \cite{huang2013sentiment,camacho-collados-etal-2022-tweetnlp}, there is a notable scarcity of labelled data for social media content, particularly in languages other than English \cite{selvaperumal2014short}; while a lot of the available datasets offer a limited taxonomy of topics \cite{vadivukarassi2019comparison}. %\jose{There is also} research that utilises semi-supervised approaches \cite{setiawan2016feature} or weakly labelled datasets \cite{yuan2018incorporating} to address the scarcity problem. 
Multilingual and cross-lingual topic classification in social media is \jose{therefore a limited explored area}. It involves dealing with content in multiple languages, addressing language-specific nuances, and ensuring effective classification. Few resources and models are designed to handle multilingual topic classification. Existing datasets e.g. in Portuguese \cite{daouadi2021optimizing}, Spanish \cite{imran2016twitter}, Urdu \cite{kausar2021hashcat} and others \cite{chowdhury2020cross}, often suffer from weak labelling or a limited taxonomy of topics, \jose{or} they are created to solve specific problems e.g. sentiment analysis \cite{muhammad2023semeval} and hate speech \cite{ousidhoum-etal-2019-multilingual}. %politics, natural disasters
This presents a gap in the field as many social media platforms have a global user base. Our work addresses this gap by introducing the {\DATASET} dataset, which includes tweets in four different languages (English, Spanish, Japanese, and Greek), thereby expanding resources for multilingual topic classification in social media.

\section{\DATASET, a Multilingual Tweet Topic Classification Benchmark}

In this section, we describe our methodology to construct, a multilingual tweet topic classification benchmark. First, we describe the original English-based TweetTopic dataset, which we take as inspiration to construct a fully multilingual dataset.

%{\DATASET}\footnote{{\DATASET} will be openly released upon acceptance.}

%\noindent{\textbf{Original TweetTopic}}
%\label{tweettopic}
TweetTopic \cite{antypas-etal-2022-twitter} \jose{is an English Twitter topic classification dataset consisting of} a total of 11,267 English tweets assigned one or more classes from a predefined list of 19 topics \jose{such as} "News \& Social Concern", "Sports", and "Fashion \& Style". The taxonomy of topics was decided by a team of social media experts and aims to cover the majority of content being shared in social media platforms. The tweets \jose{were} distributed over time, from September 2019 to October 2021 and were extracted using keywords of trending topics in each week during the period. Each entry was labelled by five different annotators, and the topic was assigned if there was an agreement of at least two annotators. 

%\subsection{\DATASET (Multilingual Extension)}

In our work, we leverage the taxonomy originally presented in TweetTopic as a foundation for collecting a 
%fresh
new set of recent tweets, leading to the introduction of {\DATASET}. {\DATASET} is \jose{mainly} distinguished by its inclusion of entries in four diverse languages: Spanish, Greek, Japanese, and English. %Specifically, we compiled, curated and annotated 1,000 new tweets for each language, amounting to a total of 4,000 tweets spanning the time frame from September 2021 to August 2022. The creation of {\DATASET} serves a dual purpose. Firstly, it aims to provide valuable resources for languages other than English and simultaneously, it extends the real-world temporal context established in the original TweetTopic, where test set instances are compiled following the date of those of the training set.

\subsection{\jose{Language Selection and} Tweet Collection}

The selection of languages was made \jose{by} taking into account their popularity and practicality. {\DATASET} is a resource that helps to the analysis of frequently used languages \jose{in X} (English, Spanish, Japanese) as well as a less frequently studied one (Greek). This linguistic diversity also provides a unique opportunity for comparative analysis between linguistically distant groups, such as Japanese and Greek. Moreover, our choice of the September 2021 to August 2022 timeframe continues the timeline of previous work and facilitates engaging in temporal analyses.

For the collection of the dataset, we follow a similar approach to that of the original TweetTopic. Initially, the \jose{Twitter} API was utilised to collect 50 tweets every two hours for each language. However, in contrast to TweetTopic, we do not use any keyword filtering in our queries. In this way, we acquire a diverse set of tweets, approximately 220,000 tweets for each language, which is closer to the real \jose{distribution of} content shared in \textit{X}.

\subsection{Preprocessing}
\label{sec:preprocessing}

Following the collection of the raw tweets we apply several preprocessing steps. First, we remove potentially remaining tweets \jose{in other languages} by using a fastText-based language identifier \cite{bojanowski-etal-2017-enriching} \jose{on top of the Twitter pre-defined language identifier}. Then, we remove tweets that are not in our target period, tweets containing incomplete sentences (too short or end in the middle of the sentence), or abusing words by \jose{applying some simple} rule-based heuristics. We also apply a \textit{near-duplication} filter to drop duplicated tweets. This process begins by normalising each tweet (i.e. remove irrelevant substrings and lemmatisation), and then retaining unique tweets only in terms of the normalised form. To ensure the quality of the tweets' content we remove entries that contain URLs, and those where multiple (more than four) emojis or mentions are present.\footnote{\dimos{Detailed number of tweets dropped in each preprocessing step can be found in Table \ref{tab:preprocessing_steps}, Appendix \ref{sec:appendix-dataset}.}} Finally, we sample 1,000 tweets from the remaining set \jose{of tweets after preprocessing} for each language. The sampling is weighted based on the retweet count of each entry as well as the follower count of the user posting the tweet. This weighting is applied with the assumption that a higher quality content is usually more popular. As a final preprocessing step we mask all mentions of non-verified users with \{USER\} to ensure the privacy of users.

\subsection{Annotation}
\label{sec:annotation}
The annotation process closely mirrored the procedure established in TweetTopic. Specifically, each entry of the dataset was annotated by five coders, where each coder had to select one or more labels from a selection of 19 topics in total. A topic was assigned to a tweet only if at least two annotators were in agreement about it. Following previous work on multi-label classification \cite{mohammad2018semeval}, we refrained from utilising a majority rule in order to create a more realistic and challenging dataset.

The coders who worked on this task were selected and filtered through the \textit{Prolific.co} platform based on their fluency in the corresponding target language. %\jose{All the annotators were compensated fairly with XXX USD per hour.} 
The actual annotation was performed through an interface created with qualtrics\textsuperscript{XM}.\footnote{The annotation guidelines for each language can be found in Appendix \ref{sec:appendix-guidelines}.} We did not utilise Amazon Mechanical Turk (AMT) due to both the lack of non-English annotators in AMT, as well as, due to the better quality of annotators present in \textit{Prolific.co}. Finally, we ensured the quality of the annotations as our research team includes native speakers in all the non-English languages, who monitored the whole annotation process for each language.

\begin{figure}
    \centering
    \includegraphics[width=\linewidth]{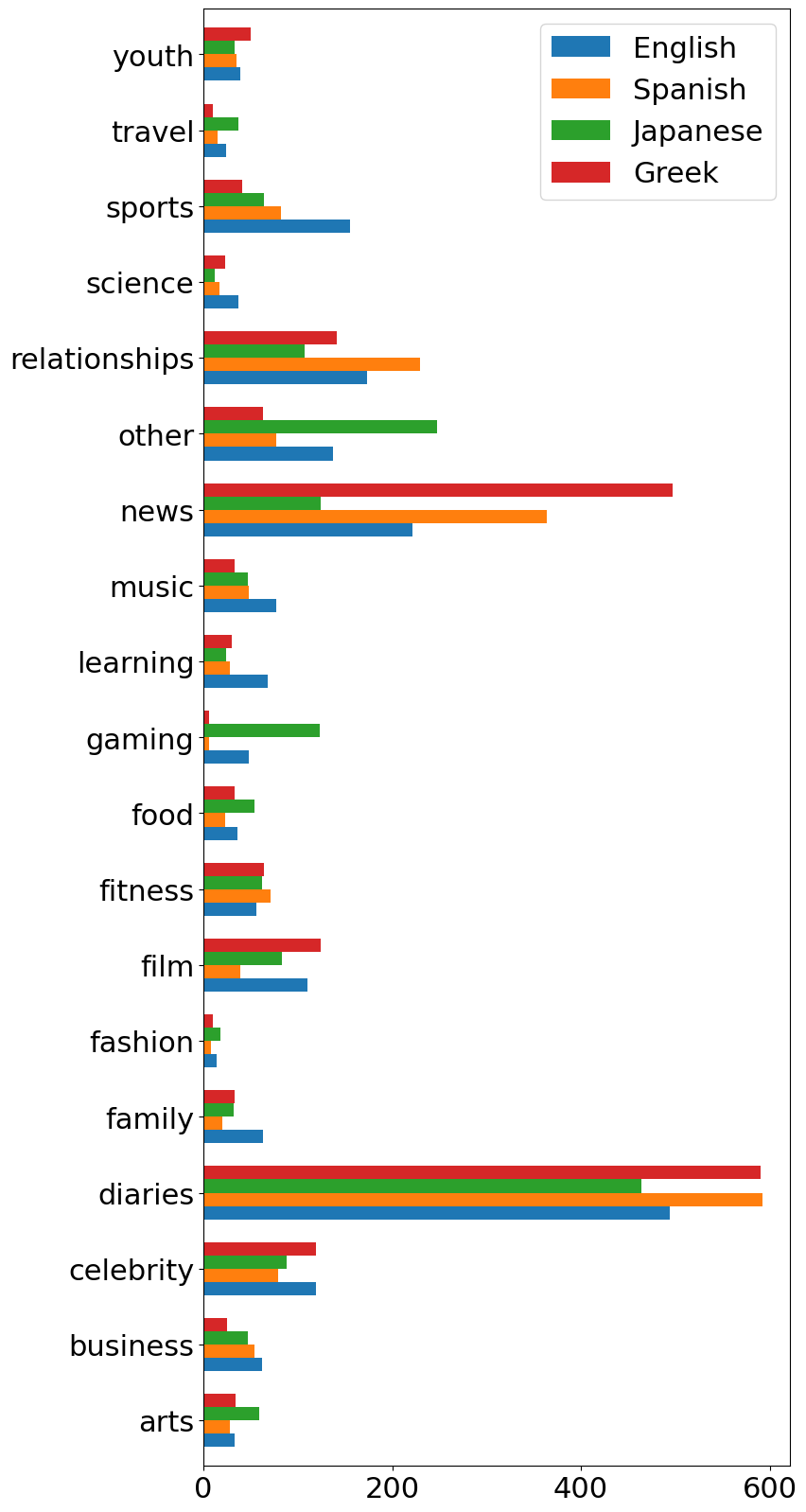}
    \caption{Number of tweets per topic and language.}
    \label{fig:topic-distribution}
\end{figure}

 %Inter-Annotator Agreement (\textit{IRA})
To assess the quality of our annotation process, we report the following three annotation agreement metrics: (1) Krippendorff's Alpha (\textit{Alpha}) \cite{krippendorff2011computing}, (2) Percent Agreement  (\textit{PA}), ratio of number of agreements to the total number of annotations, and (3) Agreement between each pair of coders on at least one label (\textit{Overlap}). When comparing our results with those achieved in the TweetTopic annotation, as presented in Table \ref{tab:annotators-agreement}, we can observe an overall smaller concordance 
 among coders. The highest \textit{Alpha} score observed was 0.26 in the Greek dataset, in contrast to TweetTopic's 0.34. Nevertheless, the agreement metrics remain on par with similar multi-label annotation tasks such as the \jose{datasets \textit{Affect in Tweets}, with a Fleiss' Kappa score of 0.26, \cite{mohammad2018semeval} and \textit{GoEmotions} \cite{demszky-etal-2020-goemotions}, with an \textit{Alpha} score of 0.24, noting} that a random annotation process would yield an \textit{Alpha} score of 0.
 %Nevertheless, the agreement metrics remain adequate when compared against similar multi-label annotation tasks \cite{mohammad2018semeval,demszky-etal-2020-goemotions}, especially when considering that a random annotation process would yield an \textit{Alpha} score of 0. Notably, the Japanese annotators encountered the greatest challenges in reaching a consensus on specific topics (\textit{Overlap}: 0.48), which may also be due to the diverging number of labels per instance across the languages.

\begin{table}[ht]
\begin{adjustbox}{width=\linewidth,center}
\begin{tabular}{l|c|c|c|c}
                    & \textbf{Alpha} & \textbf{ PA } & \textbf{Overlap}  & \textbf{AVG Topics}\\ \hline%(at least 1 class)
\textbf{English}    & 0.23                                                                   & 0.87     & 0.60  & 2.0 \\
\textbf{Spanish}    & 0.23                                                                   & 0.89     & 0.63  & 1.8\\
\textbf{Japanese}   & 0.21                                                                   & 0.87     & 0.48  & 1.7\\
\textbf{Greek}      & 0.26                                                                   & 0.89     & 0.74  & 1.9 \\ \hline\hline
\textbf{TweetTopic} & 0.34                                                                   & 0.90     & 0.70  & 1.6                                                                        
\end{tabular}
\end{adjustbox}

\caption{Annotator agreement in each language subset of {\DATASET} and TweetTopic, \jose{as well as the} average number of topics (\textit{AVG Topics}) assigned to each tweet.}
\label{tab:annotators-agreement}
\end{table}

\subsection{\jose{Descriptive Analysis}}
% avg number of topics / tweet
% en    1.966
% es    1.815
% gr    1.926
% ja    1.726
% avg number of char
% en    149.02
% es    128.93
% gr    144.71
% ja     48.58
% avg number of emojis
% en    0.431
% es    0.428
% gr    0.254
% ja    0.341
{\DATASET} encompasses a total of 361 distinct topic \jose{combinations} within its 4,000 tweets, showcasing its diversity in themes and coverage. In Table \ref{tab:examples}, we present illustrative entries from our dataset for each language, displaying \jose{various} topics. Notably, each tweet, on average, is associated with 1.8 topics, with none of the entries assigned more than 5 topics.

\paragraph{Topic overlap.} Upon examining the overlap between topics across all languages, \dimos{as depicted in Figure \ref{fig:overlap}, Appendix \ref{sec:appendix-dataset}}, we observe interesting patterns. For instance, the \textit{diaries\_\&\_daily\_life (diaries)}  topic frequently co-occurs with other topics, such as \textit{family} (79\%) and \textit{relationships} (76\%). Furthermore, there is a substantial overlap between topics %that align with expectations. a
that we expected to be closely related in online discussions.
For instance, \textit{music} and \textit{celebrities\_\&\_pop\_culture} exhibit a 45\% overlap, while \textit{youth\_\&\_student\_life (youth)} and \textit{learning\_\&\_educational (learning)} demonstrate a 25\% overlap. %These findings serve as a valuable quality assurance measure for our collected data, reinforcing the reliability of the data annotation.

\paragraph{Topic distribution.} As seen in Figure \ref{fig:topic-distribution}\footnote{A map of topic name abbreviations is provided in Appendix \ref{sec:appendix-topic_abbreviations}.}, \textit{diaries\_\&\_daily\_life} is the majority class across all four language subsets with 494, 592, 464, and 590 tweets present in English, Spanish, Japanese, and Greek respectively. When looking at less popular topics, \jose{differences} between languages start becoming apparent with \textit{news\_\&\_social\_concern} being the second most popular topic for English, Spanish, and Greek (221, 364, and 497 tweets respectively), and \textit{other\_hobbies} being the second most popular topic in Japanese (248 tweets). This is in contrast to the TweetTopic dataset which also exhibits an imbalanced distribution but to a lesser degree. This difference can be explained by the fact that in {\DATASET} we randomly extract tweets from \jose{X}, aiming to replicate a realistic distribution, rather than utilising trending keywords. These variations in the topic distributions \jose{among} the four languages, along with differences in the average post length (average number of characters: en: 149.02, es: 128.93, gr: 144.71, ja: 48.58) and the usage of emojis (average number of emojis: en: 0.43, es: 0.42, gr: 0.25, ja: 0.34), provide initial evidence of deeper differences between languages and \jose{cultures}, \jose{present initial evidence} into the challenges \jose{for} developing cross-/multi-lingual models.

\section{Experimental Setting}
In this section, we introduce the models that we evaluate using {\DATASET} and outline the various settings employed for our analysis.

\subsection{Data \& Settings}
\label{sec:settings}
To investigate the robustness of our models and the quality of the collected data, we \jose{perform a multi-purpose evaluation} in a cross-validation setting. For each language subset of {\DATASET}, we implement a 5-fold cross-validation approach, with each fold encompassing 720/80/200 tweets for the train/validation/test sets. We ensure, whenever possible, that at least one instance of each topic is represented in each split. \jose{Then, we evaluate the following settings in the test splits of {\DATASET}.} %Finally, we also utilise \jose{the existing English-language} TweetTopic \jose{as training data} and the aforementioned data splits of {\DATASET} to assess the performance of current models in cross-lingual and multilingual settings, \jose{as outlined below.}

\noindent{\textbf{Zero-shot (zero).}} No training data are provided. This setting aims to investigate the performance of zero-shot and unsupervised systems such as recent instruction tuning \cite{chung2022scaling} and generative language models \cite{bubeck2023sparks} in low-resource settings.

\noindent{\textbf{Few-shot (few).}} Five entries selected from the validation set of each fold are provided as examples. We aimed to maximise the coverage of topics present when selecting the entries. The goal of this setting is to assess the model's ability to generalise to new tasks or domains with limited training examples. For both the zero and few-shot settings the prompts utilised are similar to the ones used for the training of the BLOOMz and MT0 models in  \citet{muennighoff2022crosslingual} (see Appendix \ref{sec:appendix-prompts}).

\noindent{\textbf{Cross-lingual (TweetTopic).}} In this setting, we utilise the full English TweetTopic dataset \cite{antypas-etal-2022-twitter} \jose{as training set}. The goal of the setting is to develop a cross-lingual classifier which will be evaluated on the language-specific test sets of {\DATASET}. \jose{This setting can serve as an indication of the performance in other languages not included in {\DATASET} for which training data is not available.} In addition to the cross-lingual challenge, this setting will have the added temporal challenge, as training and test sets come from different time periods.

\noindent{\textbf{Monolingual (target).}} For each \jose{target} language, we only make use of its respective training/validation splits in each fold to fine-tune classifiers, which are then evaluated on their respective test sets of the same language. The purpose of this configuration is to assess the capabilities of classifiers across languages as well as to learn from a limited amount of data.

\noindent{\textbf{Multilingual (all \jose{languages}).}} In this scenario, we fine-tune a single model utilising all available training data in {\DATASET} in each fold, aiming to investigate the potential benefits of using a larger amount of training data and the model's capabilities in learning from \jose{labeled data in} different languages.

%\noindent{\textbf{Multilingual+ (TweetTopic + all targets)}} In each fold both TweetTopic and the training/validation splits of {\DATASET} (800 tweets for each language, 3,200 in total) are used as training/validation data while the models are evaluated on the test sets of {\DATASET}. In this setting, we combine all of the available training data for the task. %and thus we expect to see the best performing models.

\jose{For both the monolingual and multilingual settings above, we also explored the setting in which we add the original English TweetTopic as additional training data. The reason for this is to have a setting that includes all training data available, which is a common setting in many NLP tasks in which a larger amount of English data is available.}

\subsection{Comparison Models}
\label{sec:models}

We consider two types of models depending on whether they are fine-tuned, or used out of the box in zero- or few-shot settings via prompting.

\subsubsection{Fine-tuning}
We consider five different \jose{multilingual} models, \jose{both general-purpose and specialised on social media and of different sizes,} for the fine-tuning setting. %The choice of these models was determined by: (1) their multi-/cross-lingual capabilities; (2) their architecture and size; and (3) their training process. %More specifically we utilise:

\noindent \textbf{bernice} \cite{delucia-etal-2022-bernice}, a RoBERTa-based model trained on a large corpus of 2.5 billion tweets employing a customised tweet-focused tokenizer. Its training data includes 66 different languages with English, Spanish, and Japanese being the first, second, and fourth most frequent languages, making it an ideal candidate for the task at hand.

\noindent  \textbf{XLM-R (xlmr)} \cite{DBLP:journals/corr/abs-1911-02116}, \jose{a RoBERTa-like model} trained on the CommonCrawls corpus \cite{wenzek-etal-2020-ccnet} on 100 languages; and \textbf{XLM-T (xlmt)} \cite{barbieri-etal-2022-xlm}, \jose{another} XLM-R based model that utilises the last XLM-R checkpoint and further trains on a diverse dataset of over 1 billion tweets spanning over 30 languages.

% \noindent  \textbf{XLM-V (xlmv)} \cite{2023arXiv230110472L}, a multilingual model which builds upon the XLM-R architecture and introduces new tokenization approach aiming to build a more efficient vocabulary with better coverage for each language than it's predecessor.

For models based on XLM-R, we evaluate both the base and large versions. The inclusion of non-social media specific models (\textit{xlmr}) is valuable as it offers insights into their performance in scenarios where the model is not specifically trained on social media content, shedding light on the inherent challenges of such settings. The implementation provided by Hugging Face \cite{wolf-etal-2020-transformers} is used for the fine-tuning of all the models. Hyper-parameter tuning, including batch size, epochs number, learning rate, and weight decay is conducted using Ray Tune \cite{liaw2018tune}\footnote{Details of the models used can be found in Appendix \ref{sec:appendix-models}.}.

\subsubsection{Zero and Few-shot}
In order to assess the zero/few-shot capabilities of large language models in our task, we compare four models of different sizes and architectures. %Specifically we test: 

\noindent \textbf{BLOOMZ (bloomz)} \cite{muennighoff2022crosslingual}, a decoder-only model based on the BLOOM models and trained with the xP3 dataset \cite{scao2022bloom} with 7 billion parameters.

\noindent  \textbf{mt0} \cite{muennighoff2022crosslingual}, a multilingual variant of the multilingual Text-to-Text Transfer Transformer model \cite{xue2020mt5}. %trained on the mC4 corpus \cite{2019t5}, which encompasses more than 101 languages. 
\textit{Mt0}, similarly to \textit{bloomz}, is further trained on the xP3 dataset using multitask prompted finetuning.% The 13 billion parameters version of the model is used.

\noindent  \textbf{chat-gpt-3.5-turbo (chat-gpt)} from OpenaAI, \footnote{\url{https://openai.com/chatgpt}} an encoder/decoder model with approximately 175 billion parameters \cite{brown2020language}.

\noindent  \dimos{\textbf{gpt-4o } the latest and best performing model from OpenAI which significantly outperforms its predecessors. 
}

{
\renewcommand{\arraystretch}{0.991}
\begin{table*}[!h]
\centering
\resizebox{\linewidth}{!}{%
\addtolength{\tabcolsep}{-0.2em}
\begin{tabular}{cc|lccccc|ccccc|ccccc|ccccc}
\multicolumn{2}{c|}{\multirow{4}{*}{\textbf{}}} & \multicolumn{1}{c}{\textbf{}} & \multicolumn{5}{c|}{\textbf{English}} & \multicolumn{5}{c|}{\textbf{Spanish}} & \multicolumn{5}{c|}{\textbf{Japanese}} & \multicolumn{5}{c}{\textbf{Greek}} \\
\multicolumn{2}{c|}{} & \textbf{TweetTopic} & \checkmark &  &  & \checkmark & \checkmark & \checkmark &  &  & \checkmark & \checkmark & \checkmark &  &  & \checkmark & \checkmark & \checkmark &  &  & \checkmark & \checkmark \\
\multicolumn{2}{c|}{} & \textbf{Target} & \textbf{} & \textbf{\checkmark} & \textbf{} & \textbf{\checkmark} & \textbf{} & \textbf{} & \textbf{\checkmark} & \textbf{} & \textbf{\checkmark} & \textbf{} & \textbf{} & \textbf{\checkmark} & \textbf{} & \textbf{\checkmark} & \textbf{} & \textbf{} & \textbf{\checkmark} & \textbf{} & \textbf{\checkmark} & \textbf{} \\
\multicolumn{2}{c|}{} & \textbf{All} &  &  & \checkmark &  & \checkmark &  &  & \checkmark &  & \checkmark &  &  & \checkmark &  & \checkmark &  &  & \checkmark &  & \checkmark \\ \hline
\multirow{13}{*}{\rotatebox{90}{macro-F1}} & \multirow{5}{*}{\rotatebox{90}{finetuned}} & bernice & 55.9 & 42.7 & 55.2 & 58.7 & 60.3 & 52.0 & 26.6 & 51.5 & 55.8 & 55.9 & 45.8 & 39.9 & 55.2 & 53.3 & 54.3 & 41.4 & 26.4 & 40.1 & 43.4 & 44.0 \\
 &  & xlmr\_base & 47.0 & 25.1 & 45.9 & 58.0 & 57.6 & 42.4 & 11.6 & 35.1 & 48.4 & 49.1 & 34.4 & 2.7 & 39.9 & 50.1 & 52.5 & 29.5 & 12.3 & 34.2 & 40.0 & 39.7 \\
 &  & xlmr\_large & 57.2 & 51.1 & 58.7 & 60.8 & \textbf{\underline63.3} & 51.8 & 32.6 & 49.4 & 53.0 & 57.2 & 49.1 & 38.5 & 55.9 & 56.6 & 56.7 & 44.0 & 26.7 & 45.6 & 45.5 & 46.2 \\
 &  & xlmt\_base & 55.4 & 42.7 & 55.1 & 59.1 & 60.3 & 48.5 & 29.9 & 49.1 & 52.8 & 54.2 & 47.8 & 29.5 & 50.8 & 53.1 & 54.4 & 32.6 & 21.8 & 39.6 & 41.3 & 45.4 \\
 &  & xlmt\_large & 60.2 & 52.0 & 59.9 & 62.1 & 61.7 & 52.9 & 45.4 & 54.4 & 56.6 & \textbf{\underline60.0} & 50.9 & 50.9 & 57.3 & 57.2 & \textbf{\underline58.5} & 40.6 & 30.1 & 49.3 & 48.6 & 50.3 \\ \cline{2-23} 
 & \multirow{4}{*}{\rotatebox{90}{zero}} & bloomz & \multicolumn{5}{c|}{23.4} & \multicolumn{5}{c|}{15.5} & \multicolumn{5}{c|}{15.2} & \multicolumn{5}{c}{1.5} \\
 &  & mt0 & \multicolumn{5}{c|}{34.7} & \multicolumn{5}{c|}{29.2} & \multicolumn{5}{c|}{37.3} & \multicolumn{5}{c}{24.7} \\
 &  & chat-gpt & \multicolumn{5}{c|}{44.9} & \multicolumn{5}{c|}{37.2} & \multicolumn{5}{c|}{35.6} & \multicolumn{5}{c}{33.2} \\
 &  & gpt-4o & \multicolumn{5}{c|}{59.1} & \multicolumn{5}{c|}{52.4} & \multicolumn{5}{c|}{51.9} & \multicolumn{5}{c}{49.5} \\ \cline{2-23} 
 & \multirow{4}{*}{\rotatebox{90}{few}} & bloomz & \multicolumn{5}{c|}{21.0} & \multicolumn{5}{c|}{17.3} & \multicolumn{5}{c|}{14.0} & \multicolumn{5}{c}{5.2} \\
 &  & mt0 & \multicolumn{5}{c|}{35.7} & \multicolumn{5}{c|}{29.1} & \multicolumn{5}{c|}{39.0} & \multicolumn{5}{c}{25.1} \\
 &  & chat-gpt & \multicolumn{5}{c|}{54.1} & \multicolumn{5}{c|}{43.6} & \multicolumn{5}{c|}{43.9} & \multicolumn{5}{c}{39.5} \\
 &  & gpt-4o & \multicolumn{5}{c|}{60.0} & \multicolumn{5}{c|}{52.8} & \multicolumn{5}{c|}{53.3} & \multicolumn{5}{c}{\textbf{51.0}} \\ \hline
\multirow{13}{*}{\rotatebox{90}{micro-F1}} & \multirow{5}{*}{\rotatebox{90}{finetuned}} & bernice & 63.5 & 63.1 & 67.6 & 67.1 & 66.8 & 64.9 & 68.2 & 71.8 & 72.5 & 72.4 & 52.8 & 55.3 & 59.7 & 59.9 & 59.3 & 64.4 & 68.6 & 71.1 & 71.9 & 70.8 \\
 &  & xlmr\_base & 57.3 & 51.9 & 62.6 & 65.0 & 64.0 & 59.5 & 57.8 & 68.5 & 69.4 & 70.3 & 43.8 & 20.1 & 52.7 & 55.8 & 56.4 & 53.8 & 60.3 & 68.1 & 69.0 & 67.8 \\
 &  & xlmr\_large & 64.4 & 66.2 & 67.5 & 67.2 & \textbf{68.8} & 65.4 & 69.2 & 71.6 & 71.7 & 72.4 & 52.3 & 52.5 & 59.6 & 59.2 & 58.6 & 64.4 & 68.7 & 72.6 & 72.1 & 71.2 \\
 &  & xlmt\_base & 63.5 & 63.5 & 66.6 & 66.9 & 66.2 & 63.3 & 68.7 & 71.7 & 72.5 & 71.5 & 51.8 & 49.5 & 57.8 & 57.5 & 58.7 & 58.5 & 67.0 & 70.0 & 69.8 & 70.1 \\
 &  & xlmt\_large & 66.3 & 66.3 & 68.7 & \textbf{68.8} & 67.8 & 67.0 & 72.5 & 73.9 & 73.9 & \textbf{\underline74.5} & 56.0 & 59.6 & \textbf{\underline61.4} & 60.5 & 61.3 & 65.8 & 70.6 & \textbf{\underline74.5} & 73.0 & 73.4 \\ \cline{2-23} 
 & \multirow{4}{*}{\rotatebox{90}{zero}} & bloomz & \multicolumn{5}{c|}{24.3} & \multicolumn{5}{c|}{15.2} & \multicolumn{5}{c|}{19.3} & \multicolumn{5}{c}{0.7} \\
 &  & mt0 & \multicolumn{5}{c|}{38.7} & \multicolumn{5}{c|}{24.7} & \multicolumn{5}{c|}{42.7} & \multicolumn{5}{c}{43.2} \\
 &  & chat-gpt & \multicolumn{5}{c|}{48.6} & \multicolumn{5}{c|}{49.8} & \multicolumn{5}{c|}{39.2} & \multicolumn{5}{c}{46.6} \\
 &  & gpt-4o & \multicolumn{5}{c|}{63.6} & \multicolumn{5}{c|}{65.6} & \multicolumn{5}{c|}{56.6} & \multicolumn{5}{c}{65.1} \\ \cline{2-23} 
 & \multirow{4}{*}{\rotatebox{90}{few}} & bloomz & \multicolumn{5}{c|}{23.5} & \multicolumn{5}{c|}{14.6} & \multicolumn{5}{c|}{17.4} & \multicolumn{5}{c}{4.4} \\
 &  & mt0 & \multicolumn{5}{c|}{38.8} & \multicolumn{5}{c|}{25.2} & \multicolumn{5}{c|}{41.8} & \multicolumn{5}{c}{45.5} \\
 &  & chat-gpt & \multicolumn{5}{c|}{57.2} & \multicolumn{5}{c|}{54.9} & \multicolumn{5}{c|}{44.3} & \multicolumn{5}{c}{53.9} \\
 &  & gpt-4o & \multicolumn{5}{c|}{63.2} & \multicolumn{5}{c|}{62.3} & \multicolumn{5}{c|}{57.8} & \multicolumn{5}{c}{68.6}
\end{tabular}
}
\caption{F1 scores (macro \& micro average) for each setting tested in 5-fold cross validation. \jose{Fine-tuned models are evaluated on different settings depending on the used training data.} \textit{TweetTopic:}  TweetTopic was used for training; \textit{Target:} the respective language subset of {\DATASET} was used for training; \textit{All:} all language subsets of {\DATASET} were used. The best result for each language is bolded, and underlined scores indicate statistically significant difference with respect to the second best score.}
\label{tab:results_overall}
\end{table*}
}

\subsection{Evaluation Metrics}
Due to the nature of {\DATASET}, we use the macro-F1 score, which assigns equal weights to each label, as the evaluation metric. This metric is often used for multi-label classification tasks \cite{tujnas.v6i6.1329, lipton-mldiscovery, mohammad2018semeval}. In order to better understand the performance of the models and due to the imbalanced nature, which can be a challenge for a model's performance evaluation \cite{tkde.2008.239}, micro-F1 is also reported.

\section{Analysis of Results}
\label{sec:results}
The average macro and micro F1 scores for each model tested across various settings are presented in Table \ref{tab:results_overall}. %The settings include multilingual, cross-lingual, zero- and few-shot (see Section \ref{sec:settings}). 
Overall, the task presents a challenge for the tested models, with the top-performing classifier, \textit{xlmt-large}, achieving an average performance of 57.6\% macro-F1 when trained on all available data (TweetTopic and {\DATASET}). The majority of models demonstrate better micro-F1 scores, as they are not penalised as heavily for errors in less frequent topics. 

%Below we discuss the results based on the settings, models, and type of training used and identify trends that emerge.

\subsection{Setting Comparison}

\noindent{\textbf{Cross-lingual capabilities.}} 
We analyse the cross-lingual capabilities by comparing the performance of models trained exclusively on \textit{TweetTopic} with those trained solely on \textit{Target}, taking only Spanish, Japanese and Greek into consideration. A distinct pattern emerges where cross-lingual models \jose{perform competitively (a macro-F1 score of 51.1 for the best model \textit{xlmt\_large} on average)} consistently outperform their mono-lingual counterparts. For instance, the \textit{xlmr\_base} model shows a performance drop of up to 31 points in macro-F1 when tested on Japanese. On average, mono-lingual models display a performance decline of approximately 15 points when compared to their cross-lingual variants. \jose{This result is encouraging as it means that cross-lingual models may be used in languages for which training data is currently not available.} Even though the models' cross-lingual capabilities are \jose{remarkable}, it is worth noting that the smaller size of training data available on \textit{Target} \jose{(800 instances compared to the 11,267 instances in TweetTopic)} has a \jose{positive} effect on their performance. %This becomes evident when comparing models trained on \textit{TweetTopic} and \textit{Target}, and subsequently tested on English, where \jose{there is an} overall decline in performance. 

\noindent{\textbf{Multilingual vs Monolingual.}} 
The experiments reveal a consistent increase in performance for multilingual models trained on the entire {\DATASET} compared to their monolingual counterparts. On average, multilingual models achieve a 17-point improvement in macro-F1. The most significant performance boost is observed in non-English languages, with an average macro-F1 increase of approximately 18 points for Spanish, Japanese, and Greek, compared to only 12 points for the English subset. In general, we observe that cross-lingual models tend to improve as more languages are added. Performance consistently increases with the inclusion of additional target language data or by incorporating more languages. \dimos{The this trend can bee seen clearly when looking at the overall best-performing model \textit{xlmt\_large}, Figure \ref{fig:best_f1}, Appendix \ref{sec:appendix-results}.} 

%This trend can be seen more clearly in Figure \ref{fig:best_f1}, which displays the scores achieved by the overall best-performing model, \textit{xlm\_t-large}, in each language and setting.

%Furthermore, it is worth mentioning that  models trained on the entire {\DATASET} models achieve similar, or better performance, (with an average difference of 2.3 points in macro-F1) compared to those trained solely on TweetTopic, despite the latter being nearly three times larger. This trend becomes even clearer when considering languages absent in TweetTopic, where models perform even better.

\noindent{\textbf{Zero- and Few-Shot.}}
In both zero- and few-shot settings, %\textit{bloomz} and \textit{chat-gpt} 
\dimos{when considering macro-F1, \textit{bloomz}, \textit{chat-gpt}, and \textit{gtp-4o}} perform better in English and display a noticeable decline in other languages. 
%In general, \textit{chat-gpt} consistently surpasses the smaller \textit{bloomz7b} \jose{and \textit{mt0} across most languages and metrics}. An exception to this trend is \textit{mt0}, which achieves the highest performance when tested in Japanese, recording macro-F1 scores of 37.3 and 39 in zero- and few-shot settings, respectively, even surpassing \textit{chat-gpt} in zero-shot. %The performance disparity observed with \textit{mt0} on Japanese may be attributed to the prompts utilised, which are similar to those used during the original training of the model.
\dimos{In general, \textit{gpt-4o} consistently surprasses the smaller \textit{bloomz7b} and \textit{mt0}, and it's predecessor \textit{chat-gpt}, across all language and metrics.}
%The limitations of the smaller models become \jose{more apparent, however,} when looking at the minimal increase in performance in the \textit{few-shot} setting. Unlike \textit{chat-gpt}, which shows an average performance increase of 8 points in macro-F1, \textit{mt0} and \textit{bloomz} display only slight improvements compared to their \textit{zero-shot} benchmarks. In some cases, performance even decreases with \textit{bloomz} experiencing a drop of 2.4 points in macro-F1 when tested in English. \jose{This shows a better adaptability of a model such as \textit{chat-gpt} to in-context learning settings.}
\dimos{
It is interesting to note the differences in performance tha arise in the zero and few-shot benchmarks. The performance of most models, according to macro-F1, increase in the \textit{few-shot} benchmark, \textit{bloomz} being an exception and experiencing a drop of 2.4 points when tested in English. In contrast, \textit{gpt-4o} displays a decrease in micro-F1 scores across all languages indicating a consistent difficulty in maintaining performance when handling imbalanced datasets with more frequent classes.}

\subsection{Model Comparison}

\noindent{\textbf{Training Corpora.}}
Overall, models trained on \textit{X} data, consistently outperform the generic \textit{XLMR} models. Notably, both \textit{bernice} and \textit{xlmt\_base} demonstrate superior performance compared to \textit{xlmr\_base} across all settings and languages, with an average increase in macro-F1 of 11.7 and 8.3 points, respectively. This trend also appears in the larger versions, where \textit{xlmt\_large} surpasses \textit{xlmr\_large} by an average of 3 macro-F1 points across settings. The performance gap between specific \textit{X} models and generic \textit{XLMR} models widens in settings with limited training data (trained only on \textit{Target}). Specifically, the \textit{X}-specific models outperform the generic ones by a significant margin, reaching up to a 37-point increase in macro-F1 (e.g., \textit{bernice} trained on Japanese only) for the base versions and a 12-point increase for the larger versions (e.g., \textit{xlmt\_large} trained on Spanish only). %Finally, it is worth mentioning that despite \textit{bernice} not being based on an architecture designed for cross- or multilingual tasks, it consistently outperforms the similarly-sized \textit{xlmt\_base} in almost every setting. This discrepancy could potentially be attributed to differences in the size of their original training corpora. 
\jose{These} results highlight the \jose{benefit of} training models on specific domain data. %and emphasise the impact of training data availability on model performance. 

\begin{table}[]
\centering
\scalebox{1}{
\begin{tabular}{l|c|c}
\textbf{LN} & \textbf{xlmt\_large} & \textbf{gpt-4o} \\ \hline
en & learning, 78 & other, 85 \\
 & arts, 76 & learning, 73 \\
 & other, 74 & youth, 69 \\ \hline
ja & news, 66 & business, 84 \\
 & business, 64 & arts, 76 \\
 & arts, 59 & relationships, 74 \\ \hline
es & other, 83 & other, 82 \\
 & arts, 68 & youth, 80 \\
 & travel, 67 & business, 75 \\ \hline
gr & other, 89 & other, 95 \\
 & youth, 86 & youth, 87 \\
 & arts, 76 & science, 71
\end{tabular}
}
\caption{\dimos{Topics with the highest occurrences of False Negatives errors (topic, error \%). The results of \textit{xlmt-large} when trained on \textit{TweetTopic} and \textit{All}, and of \textit{gpt-4o} in the \textit{few-shot} setting are displayed.}}
\label{tab:top-errors}
\end{table}

\noindent{\textbf{Fine-tuned models vs few-shot LLMs.}} 
The experimental results of LLMs reveal that the task is challenging even for larger models. When compared to the finetuned models, the best performing LLM, \dimos{\textit{gpt-4o} in the \textit{few-shot} setting, achieves comparable results with \textit{xlmt\_base}  when fine-tuned  on all available datasets, with average macro-F1 of 54.3 and 53.6 for \textit{gpt-4o} and \textit{xlmt\_base} respectively, however it achieves the best macro-F1 performance in Greek across all models. In order to better understand the behaviour of each type of model, Table \ref{tab:recall_scores} displays the average macro Recall and Precision scores achieved by four models of different architectures. Notably, \textit{chat-gpt} seems to struggle more with identifying correctly the assigned labels, as it achieves relatively smaller Precision scores compared to other models. Instead, recall values of \textit{chat-gpt} are similar or higher than other models, particularly for English and Spanish. On average, \textit{chat-gpt} predicts 2, 2.5, 1.5, and 1.4 labels per tweet in English, Spanish, Japanese and Greek, respectively. In contrast, the best performing finetuned model, \textit{xlmt\_large}, predicts a more consistent average of 1.7, 1.7, 1.7, and 1.8 labels per tweet on the same languages.} %This behaviour suggests that if the problem demands high recall LLMs can be the preferable choice to use.}

\subsection{Error Analysis}
\dimos{Using the best overall performing models, \textit{xlmt-large} trained on \textit{TweetTopic} and \textit{All languages}, and \textit{gpt-4o} in a \textit{few-shot} setting, we attempt to identify patterns in the topics which it struggles the most. Generally, both models attain relatively low recall values (Table \ref{tab:recall_scores}) compared to precision. We analyse this behaviour by examining the topics with the highest occurrences of errors by analysing the False Negative rates (Table \ref{tab:top-errors}). It is interesting to note the high occurrences of errors noted on the \textit{xlmt\_large} results across all languages within the relatively infrequent \textit{Arts \& Culture} topic, with error rates of 76\%, 59\%, 68\%, and 76\% for English, Japanese, Spanish, and Greek, respectively. In contrast, \textit{gpt-4o} appears to struggle more with the \textit{Youth \& Student Life} topic.}

\dimos{Investigating the models' performance in more detail (Tables \ref{tab:xlmt-details} and \ref{tab:gpt-4o-details}, Appendix \ref{sec:appendix-models}), reveals a significant weaknesses for both \textit{xlmt\_large} and \textit{gpt-4o} in the \textit{Other Hobbies} category. Both models exhibit low performance in all languages with \textit{xlmt\_large}  and \textit{gpt-4o} achieving 28\% and 25\% average F1 respectively,  highlighting the difficulty in classifying diverse and less defined subjects. }
% Using th'e best overall performing model, \textit{xlmt-large} trained on \textit{TweetTopic} and \textit{All languages}, we attempt to identify patterns in the topics which it struggles the most. Generally, the model attains \jose{relatively} low recall values (Table \ref{tab:recall_scores}) \jose{compared to precision}. We \jose{analyse this behaviour by examining} the topics with the highest occurrences of errors by analysing the False Negative rates (Table \ref{tab:top-errors}). High occurrences of errors are noted across all languages within the \jose{relatively infrequent} \textit{Arts \& Culture} topic, with error rates of 76\%, 59\%, 68\%, and 76\% for English, Japanese, Spanish, and Greek, respectively, \jose{as well as for the cross-cutting \textit{Other Hobbies} topic.}

\dimos{When looking at examples where the models tend to struggle more, there are clear errors like the tweet `\textit{Being on the other side of the casting table today was so much fun. Saying "just have fun with it" and seeing actors literally just have fun with it was amazin}` being classified by \textit{gpt-4o} as "Family" but also there are entries such as \textit{"what are the best web3/crypto newsletters out there not many people know about?"} which is labelled as "News \& Social Concern", "Science \& Technology" by \textit{xlmt\_large} instead of "News \& Social Concern", "Business \& Entrepreneurs", an arguably valid classification. This behaviour illustrates the difficulty of the task for both human annotators and language models.
}

%It is interesting to note that \textit{chat-gpt} seems to struggle more with identifying correctly the assigned labels, as it achieves relatively smaller Precision scores compared to other models. Table \ref{tab:recall_scores} displays the average macro Recall and Precision scores achieved by four models of different architectures. Recall values of \textit{chat-gpt} are \jose{similar or} higher than its competition. 

%When looking at examples where OpenAI's model assigns more labels, there are clear errors like the tweet `\textit{Being on the other side of the casting table today was so much fun. Saying "just have fun with it" and seeing actors literally just have fun with it was amazin}` being classified as "Family" but also \jose{there are} entries \jose{such as} \textit{"what are the best web3/crypto newsletters out there not many people know about?"} which is labelled as "News \& Social Concern", "Science \& Technology" by \textit{chat-gpt} instead of "News \& Social Concern", "Business \& Entrepreneurs", an arguably valid classification. This behaviour illustrates the difficulty of the task for both human annotators and language models.

\begin{table}[]
\begin{adjustbox}{width=\linewidth,center}
\begin{tabular}{l|cccc|cccc}
 & \multicolumn{4}{c|}{\textbf{Precision}} & \multicolumn{4}{c}{\textbf{Recall}} \\
 & En & Es & Ja & Gr & En & Es & Ja & Gr \\ \hline
chat-gpt & 53.0 & 39.5 & 46.5 & 44.0 & \textbf{63.4} & \textbf{63.0} & 49.6 & 43.0 \\
gpt-4o & 67.6  & 61.2  &  60.8 &  \textbf{63.0} & 58.2 & 53.4  & 52.6  & 47.6  \\ \hline
bernice & 65.9 & 61.9 & 57.6 & 50.0 & 58.8 & 56.3 & 54.5 & 43.1 \\
xlm\_t & \textbf{69.2} & \textbf{67.7} & \textbf{62.1} & 61.1 & 58.1 & 57.9 & \textbf{58.4} & \textbf{48.2}
\end{tabular}
\end{adjustbox}
\caption{Average macro Precision and Recall scores. Results from the few-shot setting are considered for \textit{chat-gpt} and \textit{gpt-4o}. For the \textit{bernice} and \textit{xlm\_t} results we considered models trained on TweetTopic and {\DATASET} }
\label{tab:recall_scores}
\end{table}

\section{Conclusions}

\jose{The aim of this paper is to expand} the resources available for the task of tweet classification, particularly in a multi-label setting and across multiple languages. We introduce the new {\DATASET} dataset, which includes tweets in English, Spanish, Japanese, and Greek, and is centred around a taxonomy of 19 social media topics. This dataset addresses the lack of labelled multilingual \textit{X} data and encourages the development of new methods for multilingual topic classification.

We explore different model architectures and experimental settings, including zero-shot, monolingual, cross-lingual, and multilingual approaches, to tackle the challenge of multilingual topic classification in social media. Our findings indicate that the task is challenging, especially for less-resourced languages, and that models perform better when trained on a combination of \jose{data in various languages}. \jose{Importantly, our analysis shows how recent LLMs underperform in few-shot settings in comparison to more efficient but fully-trained multilingual masked language models.} Further research \jose{should} focus on addressing these challenges and enhancing the performance of models in a cross-lingual and multilingual context, \jose{for which {\DATASET} can contribute to as a reliable benchmark.}

%{\DATASET} can be a valuable resource for researchers and industry experts dealing with real-world applications in social media. It is a valuable tool that can help the development of multilingual solutions and encourages the exploration of new approaches. This dataset addresses the complexities inherent in the diffusion of information and news in the realm of social media. The results also highlight the need for more comprehensive datasets and improved models for languages with limited resources. Further research can focus on addressing these challenges and enhancing the performance of models in a cross-lingual and multilingual context.

\section{Limitations}
\label{sec:limitations}
In this paper, we introduce a valuable new resource that is expected to benefit a wide range of researchers and industry professionals. It is important to acknowledge that there may be differing opinions regarding the methodology used for aggregating the data in {\DATASET}, specifically the requirement for two annotators' agreement. In any case, we plan to release all the collected annotations, along with the dataset version used in our experiments, to facilitate transparency and further research. \jose{The number of languages included in {\DATASET} selected is relatively small given budget constraints.}

\jose{Finally}, it is important to highlight that while our paper provides a comprehensive analysis of the cross-/multi-lingual capabilities of five different models, substantial research opportunities remain in exploring the potential of alternative classifiers. This includes investigating the performance \jose{and fine-tuning} of larger models, considering diverse architectures, and optimising the prompts used for one-shot and few-shot learning.

\section{Ethics Statement}
\label{sec:ethics}
We acknowledge the importance of the ACL Code of Ethics, and are committed to following the guidelines in the proposed task. Given that our task includes user generated content we are committed to respect the privacy of the users, by replacing each user mention in the texts with a placeholder. 

We also make sure to fairly treat the annotators who labelled the dataset, by 1) fairly compensating them with an average of £8 per hour; and 2) do not share or store their personal information. \dimos{Overall, the total time of annotation was approximately 180 hours with a median time of 25 minutes for each "batch" of 50 tweets and each batch requiring 5 coders.}

Finally, we acknowledge the potential concerns around the analysis of individual behaviours using our dataset, but we designed the tasks to focus on aggregated social media content, by measuring systems performances on aggregated data rather than at individual user level. 
{\DATASET} will be shared under the CC BY-NC 4.0 Deed (Attribution-NonCommercial 4.0 International).

\bibliography{anthology, custom}

\begin{thebibliography}{50}
\providecommand{\natexlab}[1]{#1}

\bibitem[{Ansari et~al.(2020)Ansari, Aziz, Siddiqui, Mehra, and Singh}]{ansari2020analysis}
Mohd~Zeeshan Ansari, Mohd-Bilal Aziz, MO~Siddiqui, H~Mehra, and KP~Singh. 2020.
\newblock Analysis of political sentiment orientations on twitter.
\newblock \emph{Procedia computer science}, 167:1821--1828.

\bibitem[{Antypas et~al.(2022)Antypas, Ushio, Camacho-Collados, Silva, Neves, and Barbieri}]{antypas-etal-2022-twitter}
Dimosthenis Antypas, Asahi Ushio, Jose Camacho-Collados, Vitor Silva, Leonardo Neves, and Francesco Barbieri. 2022.
\newblock \href {https://aclanthology.org/2022.coling-1.299} {{T}witter topic classification}.
\newblock In \emph{Proceedings of the 29th International Conference on Computational Linguistics}, pages 3386--3400, Gyeongju, Republic of Korea. International Committee on Computational Linguistics.

\bibitem[{Barbieri et~al.(2022)Barbieri, Espinosa~Anke, and Camacho-Collados}]{barbieri-etal-2022-xlm}
Francesco Barbieri, Luis Espinosa~Anke, and Jose Camacho-Collados. 2022.
\newblock \href {https://aclanthology.org/2022.lrec-1.27} {{XLM}-{T}: Multilingual language models in {T}witter for sentiment analysis and beyond}.
\newblock In \emph{Proceedings of the Thirteenth Language Resources and Evaluation Conference}, pages 258--266, Marseille, France. European Language Resources Association.

\bibitem[{Barbieri et~al.(2014)Barbieri, Saggion, and Ronzano}]{barbieri-etal-2014-modelling}
Francesco Barbieri, Horacio Saggion, and Francesco Ronzano. 2014.
\newblock \href {https://doi.org/10.3115/v1/W14-2609} {Modelling sarcasm in {T}witter, a novel approach}.
\newblock In \emph{Proceedings of the 5th Workshop on Computational Approaches to Subjectivity, Sentiment and Social Media Analysis}, pages 50--58, Baltimore, Maryland. Association for Computational Linguistics.

\bibitem[{Bianchi et~al.(2021)Bianchi, Terragni, Hovy, Nozza, and Fersini}]{bianchi-etal-2021-cross}
Federico Bianchi, Silvia Terragni, Dirk Hovy, Debora Nozza, and Elisabetta Fersini. 2021.
\newblock \href {https://doi.org/10.18653/v1/2021.eacl-main.143} {Cross-lingual contextualized topic models with zero-shot learning}.
\newblock In \emph{Proceedings of the 16th Conference of the European Chapter of the Association for Computational Linguistics: Main Volume}, pages 1676--1683, Online. Association for Computational Linguistics.

\bibitem[{Blei et~al.(2003)Blei, Ng, and Jordan}]{lda}
David~M. Blei, Andrew~Y. Ng, and Michael~I. Jordan. 2003.
\newblock Latent dirichlet allocation.
\newblock \emph{J. Mach. Learn. Res.}, 3(null):993–1022.

\bibitem[{Bojanowski et~al.(2017)Bojanowski, Grave, Joulin, and Mikolov}]{bojanowski-etal-2017-enriching}
Piotr Bojanowski, Edouard Grave, Armand Joulin, and Tomas Mikolov. 2017.
\newblock \href {https://doi.org/10.1162/tacl_a_00051} {Enriching word vectors with subword information}.
\newblock \emph{Transactions of the Association for Computational Linguistics}, 5:135--146.

\bibitem[{Brown et~al.(2020)Brown, Mann, Ryder, Subbiah, Kaplan, Dhariwal, Neelakantan, Shyam, Sastry, Askell, Agarwal, Herbert-Voss, Krueger, Henighan, Child, Ramesh, Ziegler, Wu, Winter, Hesse, Chen, Sigler, Litwin, Gray, Chess, Clark, Berner, McCandlish, Radford, Sutskever, and Amodei}]{brown2020language}
Tom~B. Brown, Benjamin Mann, Nick Ryder, Melanie Subbiah, Jared Kaplan, Prafulla Dhariwal, Arvind Neelakantan, Pranav Shyam, Girish Sastry, Amanda Askell, Sandhini Agarwal, Ariel Herbert-Voss, Gretchen Krueger, Tom Henighan, Rewon Child, Aditya Ramesh, Daniel~M. Ziegler, Jeffrey Wu, Clemens Winter, Christopher Hesse, Mark Chen, Eric Sigler, Mateusz Litwin, Scott Gray, Benjamin Chess, Jack Clark, Christopher Berner, Sam McCandlish, Alec Radford, Ilya Sutskever, and Dario Amodei. 2020.
\newblock \href {https://arxiv.org/abs/2005.14165} {Language models are few-shot learners}.
\newblock \emph{Preprint}, arXiv:2005.14165.

\bibitem[{Bubeck et~al.(2023)Bubeck, Chandrasekaran, Eldan, Gehrke, Horvitz, Kamar, Lee, Lee, Li, Lundberg et~al.}]{bubeck2023sparks}
S{\'e}bastien Bubeck, Varun Chandrasekaran, Ronen Eldan, Johannes Gehrke, Eric Horvitz, Ece Kamar, Peter Lee, Yin~Tat Lee, Yuanzhi Li, Scott Lundberg, et~al. 2023.
\newblock Sparks of artificial general intelligence: Early experiments with gpt-4.
\newblock \emph{arXiv preprint arXiv:2303.12712}.

\bibitem[{Camacho-collados et~al.(2022)Camacho-collados, Rezaee, Riahi, Ushio, Loureiro, Antypas, Boisson, Espinosa~Anke, Liu, and Mart{\'\i}nez~C{\'a}mara}]{camacho-collados-etal-2022-tweetnlp}
Jose Camacho-collados, Kiamehr Rezaee, Talayeh Riahi, Asahi Ushio, Daniel Loureiro, Dimosthenis Antypas, Joanne Boisson, Luis Espinosa~Anke, Fangyu Liu, and Eugenio Mart{\'\i}nez~C{\'a}mara. 2022.
\newblock \href {https://doi.org/10.18653/v1/2022.emnlp-demos.5} {{T}weet{NLP}: Cutting-edge natural language processing for social media}.
\newblock In \emph{Proceedings of the 2022 Conference on Empirical Methods in Natural Language Processing: System Demonstrations}, pages 38--49, Abu Dhabi, UAE. Association for Computational Linguistics.

\bibitem[{Card et~al.(2017)Card, Tan, and Smith}]{card2017neural}
Dallas Card, Chenhao Tan, and Noah~A Smith. 2017.
\newblock Neural models for documents with metadata.
\newblock \emph{arXiv preprint arXiv:1705.09296}.

\bibitem[{Chowdhury et~al.(2020)Chowdhury, Caragea, and Caragea}]{chowdhury2020cross}
Jishnu~Ray Chowdhury, Cornelia Caragea, and Doina Caragea. 2020.
\newblock Cross-lingual disaster-related multi-label tweet classification with manifold mixup.
\newblock In \emph{Proceedings of the 58th Annual Meeting of the Association for Computational Linguistics: Student Research Workshop}, pages 292--298.

\bibitem[{Chua and Banerjee(2016)}]{chua2016linguistic}
Alton~YK Chua and Snehasish Banerjee. 2016.
\newblock Linguistic predictors of rumor veracity on the internet.
\newblock In \emph{Proceedings of the International MultiConference of Engineers and Computer Scientists}, volume~1, page 387. Nanyang Technological University Singapore.

\bibitem[{Chung et~al.(2022)Chung, Hou, Longpre, Zoph, Tay, Fedus, Li, Wang, Dehghani, Brahma et~al.}]{chung2022scaling}
Hyung~Won Chung, Le~Hou, Shayne Longpre, Barret Zoph, Yi~Tay, William Fedus, Eric Li, Xuezhi Wang, Mostafa Dehghani, Siddhartha Brahma, et~al. 2022.
\newblock Scaling instruction-finetuned language models.
\newblock \emph{arXiv preprint arXiv:2210.11416}.

\bibitem[{Conneau et~al.(2019)Conneau, Khandelwal, Goyal, Chaudhary, Wenzek, Guzm{\'{a}}n, Grave, Ott, Zettlemoyer, and Stoyanov}]{DBLP:journals/corr/abs-1911-02116}
Alexis Conneau, Kartikay Khandelwal, Naman Goyal, Vishrav Chaudhary, Guillaume Wenzek, Francisco Guzm{\'{a}}n, Edouard Grave, Myle Ott, Luke Zettlemoyer, and Veselin Stoyanov. 2019.
\newblock \href {https://arxiv.org/abs/1911.02116} {Unsupervised cross-lingual representation learning at scale}.
\newblock \emph{CoRR}, abs/1911.02116.

\bibitem[{Daouadi et~al.(2021)Daouadi, Reba{\"\i}, and Amous}]{daouadi2021optimizing}
Kheir~Eddine Daouadi, Rim~Zghal Reba{\"\i}, and Ikram Amous. 2021.
\newblock Optimizing semantic deep forest for tweet topic classification.
\newblock \emph{Information Systems}, 101:101801.

\bibitem[{DeLucia et~al.(2022)DeLucia, Wu, Mueller, Aguirre, Resnik, and Dredze}]{delucia-etal-2022-bernice}
Alexandra DeLucia, Shijie Wu, Aaron Mueller, Carlos Aguirre, Philip Resnik, and Mark Dredze. 2022.
\newblock \href {https://doi.org/10.18653/v1/2022.emnlp-main.415} {Bernice: A multilingual pre-trained encoder for {T}witter}.
\newblock In \emph{Proceedings of the 2022 Conference on Empirical Methods in Natural Language Processing}, pages 6191--6205, Abu Dhabi, United Arab Emirates. Association for Computational Linguistics.

\bibitem[{Demszky et~al.(2020)Demszky, Movshovitz-Attias, Ko, Cowen, Nemade, and Ravi}]{demszky-etal-2020-goemotions}
Dorottya Demszky, Dana Movshovitz-Attias, Jeongwoo Ko, Alan Cowen, Gaurav Nemade, and Sujith Ravi. 2020.
\newblock \href {https://doi.org/10.18653/v1/2020.acl-main.372} {{G}o{E}motions: A dataset of fine-grained emotions}.
\newblock In \emph{Proceedings of the 58th Annual Meeting of the Association for Computational Linguistics}, pages 4040--4054, Online. Association for Computational Linguistics.

\bibitem[{Greene and Cunningham(2006)}]{greene2006practical}
Derek Greene and P{\'a}draig Cunningham. 2006.
\newblock Practical solutions to the problem of diagonal dominance in kernel document clustering.
\newblock In \emph{Proceedings of the 23rd international conference on Machine learning}, pages 377--384.

\bibitem[{Grootendorst(2022)}]{bertopic}
Maarten Grootendorst. 2022.
\newblock \href {https://doi.org/10.48550/ARXIV.2203.05794} {Bertopic: Neural topic modeling with a class-based tf-idf procedure}.
\newblock \emph{arXiv preprint}.

\bibitem[{Hazaa et~al.(2023)Hazaa, Ba-Alwi, and Albared}]{tujnas.v6i6.1329}
M.~A.~S. Hazaa, F.~M. Ba-Alwi, and M.~Albared. 2023.
\newblock \href {https://doi.org/10.59167/tujnas.v6i6.1329} {A proposed model for focused crawling and automatic text classification of online crime web pages}.
\newblock \emph{Thamar University Journal of Natural \& Applied Sciences}, 6:65--81.

\bibitem[{He and Garcia(2009)}]{tkde.2008.239}
H.~He and E.~Garcia. 2009.
\newblock \href {https://doi.org/10.1109/tkde.2008.239} {Learning from imbalanced data}.
\newblock \emph{IEEE Transactions on Knowledge and Data Engineering}, 21:1263--1284.

\bibitem[{Hersh et~al.(1994)Hersh, Buckley, Leone, and Hickam}]{hersh1994ohsumed}
William Hersh, Chris Buckley, TJ~Leone, and David Hickam. 1994.
\newblock Ohsumed: An interactive retrieval evaluation and new large test collection for research.
\newblock In \emph{SIGIR’94}, pages 192--201. Springer.

\bibitem[{Hu et~al.(2021)Hu, Wang, Luo, Zhang, Huang, Yan, Liu, Ly, Kacker, She et~al.}]{hu2021revealing}
Tao Hu, Siqin Wang, Wei Luo, Mengxi Zhang, Xiao Huang, Yingwei Yan, Regina Liu, Kelly Ly, Viraj Kacker, Bing She, et~al. 2021.
\newblock Revealing public opinion towards covid-19 vaccines with twitter data in the united states: spatiotemporal perspective.
\newblock \emph{Journal of Medical Internet Research}, 23(9):e30854.

\bibitem[{Huang et~al.(2013)Huang, Peng, Li, and Lee}]{huang2013sentiment}
Shu Huang, Wei Peng, Jingxuan Li, and Dongwon Lee. 2013.
\newblock Sentiment and topic analysis on social media: a multi-task multi-label classification approach.
\newblock In \emph{Proceedings of the 5th annual ACM web science conference}, pages 172--181.

\bibitem[{Imran et~al.(2016)Imran, Mitra, and Castillo}]{imran2016twitter}
Muhammad Imran, Prasenjit Mitra, and Carlos Castillo. 2016.
\newblock Twitter as a lifeline: Human-annotated twitter corpora for nlp of crisis-related messages.
\newblock \emph{arXiv preprint arXiv:1605.05894}.

\bibitem[{Kausar et~al.(2021)Kausar, Tahir, and Mehmood}]{kausar2021hashcat}
Soufia Kausar, Bilal Tahir, and Muhammad~Amir Mehmood. 2021.
\newblock Hashcat: A novel approach for the topic classification of multilingual twitter trends.
\newblock In \emph{2021 International Conference on Frontiers of Information Technology (FIT)}, pages 212--217. IEEE.

\bibitem[{Krippendorff(2011)}]{krippendorff2011computing}
Klaus Krippendorff. 2011.
\newblock Computing krippendorff's alpha-reliability.

\bibitem[{Lang(1995)}]{lang1995newsweeder}
Ken Lang. 1995.
\newblock Newsweeder: Learning to filter netnews.
\newblock In \emph{Machine Learning Proceedings 1995}, pages 331--339. Elsevier.

\bibitem[{Lazaridou et~al.(2021)Lazaridou, Kuncoro, Gribovskaya, Agrawal, Liska, Terzi, Gimenez, de~Masson~d'Autume, Kocisky, Ruder et~al.}]{lazaridou2021mind}
Angeliki Lazaridou, Adhi Kuncoro, Elena Gribovskaya, Devang Agrawal, Adam Liska, Tayfun Terzi, Mai Gimenez, Cyprien de~Masson~d'Autume, Tomas Kocisky, Sebastian Ruder, et~al. 2021.
\newblock Mind the gap: Assessing temporal generalization in neural language models.
\newblock \emph{Advances in Neural Information Processing Systems}, 34.

\bibitem[{Lewis et~al.(2004)Lewis, Yang, Russell-Rose, and Li}]{lewis2004rcv1}
David~D Lewis, Yiming Yang, Tony Russell-Rose, and Fan Li. 2004.
\newblock Rcv1: A new benchmark collection for text categorization research.
\newblock \emph{Journal of machine learning research}, 5(Apr):361--397.

\bibitem[{Liaw et~al.(2018)Liaw, Liang, Nishihara, Moritz, Gonzalez, and Stoica}]{liaw2018tune}
Richard Liaw, Eric Liang, Robert Nishihara, Philipp Moritz, Joseph~E Gonzalez, and Ion Stoica. 2018.
\newblock Tune: A research platform for distributed model selection and training.
\newblock \emph{arXiv preprint arXiv:1807.05118}.

\bibitem[{Lipton et~al.(2014)Lipton, Elkan, and Naryanaswamy}]{lipton-mldiscovery}
Z.~C. Lipton, C.~Elkan, and B.~Naryanaswamy. 2014.
\newblock \href {https://doi.org/10.1007/978-3-662-44851-9_15} {Optimal thresholding of classifiers to maximize f1 measure}.
\newblock \emph{Machine Learning and Knowledge Discovery in Databases}, pages 225--239.

\bibitem[{Mohammad et~al.(2018)Mohammad, Bravo-Marquez, Salameh, and Kiritchenko}]{mohammad2018semeval}
Saif Mohammad, Felipe Bravo-Marquez, Mohammad Salameh, and Svetlana Kiritchenko. 2018.
\newblock Semeval-2018 task 1: Affect in tweets.
\newblock In \emph{Proceedings of the 12th international workshop on semantic evaluation}, pages 1--17.

\bibitem[{Muennighoff et~al.(2022)Muennighoff, Wang, Sutawika, Roberts, Biderman, Scao, Bari, Shen, Yong, Schoelkopf et~al.}]{muennighoff2022crosslingual}
Niklas Muennighoff, Thomas Wang, Lintang Sutawika, Adam Roberts, Stella Biderman, Teven~Le Scao, M~Saiful Bari, Sheng Shen, Zheng-Xin Yong, Hailey Schoelkopf, et~al. 2022.
\newblock Crosslingual generalization through multitask finetuning.
\newblock \emph{arXiv preprint arXiv:2211.01786}.

\bibitem[{Muhammad et~al.(2023)Muhammad, Abdulmumin, Yimam, Adelani, Ahmad, Ousidhoum, Ayele, Mohammad, Beloucif, and Ruder}]{muhammad2023semeval}
Shamsuddeen~Hassan Muhammad, Idris Abdulmumin, Seid~Muhie Yimam, David~Ifeoluwa Adelani, Ibrahim~Sa’id Ahmad, Nedjma Ousidhoum, Abinew~Ali Ayele, Saif Mohammad, Meriem Beloucif, and Sebastian Ruder. 2023.
\newblock Semeval-2023 task 12: Sentiment analysis for african languages (afrisenti-semeval).
\newblock In \emph{Proceedings of the 17th International Workshop on Semantic Evaluation (SemEval-2023)}, pages 2319--2337.

\bibitem[{Ousidhoum et~al.(2019)Ousidhoum, Lin, Zhang, Song, and Yeung}]{ousidhoum-etal-2019-multilingual}
Nedjma Ousidhoum, Zizheng Lin, Hongming Zhang, Yangqiu Song, and Dit-Yan Yeung. 2019.
\newblock \href {https://doi.org/10.18653/v1/D19-1474} {Multilingual and multi-aspect hate speech analysis}.
\newblock In \emph{Proceedings of the 2019 Conference on Empirical Methods in Natural Language Processing and the 9th International Joint Conference on Natural Language Processing (EMNLP-IJCNLP)}, pages 4675--4684, Hong Kong, China. Association for Computational Linguistics.

\bibitem[{Rosen-Zvi et~al.(2004)Rosen-Zvi, Griffiths, Steyvers, and Smyth}]{rosen2004author}
Michal Rosen-Zvi, Thomas Griffiths, Mark Steyvers, and Padhraic Smyth. 2004.
\newblock The author-topic model for authors and documents.
\newblock In \emph{Proceedings of the 20th conference on Uncertainty in artificial intelligence}, pages 487--494.

\bibitem[{Scao et~al.(2022)Scao, Fan, Akiki, Pavlick, Ili{\'c}, Hesslow, Castagn{\'e}, Luccioni, Yvon, Gall{\'e} et~al.}]{scao2022bloom}
Teven~Le Scao, Angela Fan, Christopher Akiki, Ellie Pavlick, Suzana Ili{\'c}, Daniel Hesslow, Roman Castagn{\'e}, Alexandra~Sasha Luccioni, Fran{\c{c}}ois Yvon, Matthias Gall{\'e}, et~al. 2022.
\newblock Bloom: A 176b-parameter open-access multilingual language model.
\newblock \emph{arXiv preprint arXiv:2211.05100}.

\bibitem[{Schlichtkrull et~al.(2023)Schlichtkrull, Ousidhoum, and Vlachos}]{schlichtkrull2023intended}
Michael Schlichtkrull, Nedjma Ousidhoum, and Andreas Vlachos. 2023.
\newblock The intended uses of automated fact-checking artefacts: Why, how and who.
\newblock \emph{arXiv preprint arXiv:2304.14238}.

\bibitem[{Selvaperumal and Suruliandi(2014)}]{selvaperumal2014short}
P~Selvaperumal and A~Suruliandi. 2014.
\newblock A short message classification algorithm for tweet classification.
\newblock In \emph{2014 International Conference on Recent Trends in Information Technology}, pages 1--3. IEEE.

\bibitem[{Steinskog et~al.(2017)Steinskog, Therkelsen, and Gamb{\"a}ck}]{steinskog-etal-2017-twitter}
Asbj{\o}rn Steinskog, Jonas Therkelsen, and Bj{\"o}rn Gamb{\"a}ck. 2017.
\newblock \href {https://aclanthology.org/W17-0210} {{T}witter topic modeling by tweet aggregation}.
\newblock In \emph{Proceedings of the 21st Nordic Conference on Computational Linguistics}, pages 77--86, Gothenburg, Sweden. Association for Computational Linguistics.

\bibitem[{Steyvers and Griffiths(2007)}]{steyvers2007probabilistic}
Mark Steyvers and Tom Griffiths. 2007.
\newblock Probabilistic topic models.
\newblock \emph{Handbook of latent semantic analysis}, 427(7):424--440.

\bibitem[{Vadivukarassi et~al.(2019)Vadivukarassi, Puviarasan, and Aruna}]{vadivukarassi2019comparison}
M~Vadivukarassi, N~Puviarasan, and P~Aruna. 2019.
\newblock A comparison of supervised machine learning approaches for categorized tweets.
\newblock In \emph{International Conference on Intelligent Data Communication Technologies and Internet of Things (ICICI) 2018}, pages 422--430. Springer.

\bibitem[{Wang et~al.(2017)Wang, Liakata, Zubiaga, and Procter}]{10.1007/978-3-319-67256-4_30}
B.~Wang, M.~Liakata, A.~Zubiaga, and R.~Procter. 2017.
\newblock \href {https://doi.org/10.1007/978-3-319-67256-4_30} {A hierarchical topic modelling approach for tweet clustering}.
\newblock \emph{Lecture Notes in Computer Science}, pages 378--390.

\bibitem[{Wenzek et~al.(2020)Wenzek, Lachaux, Conneau, Chaudhary, Guzm{\'a}n, Joulin, and Grave}]{wenzek-etal-2020-ccnet}
Guillaume Wenzek, Marie-Anne Lachaux, Alexis Conneau, Vishrav Chaudhary, Francisco Guzm{\'a}n, Armand Joulin, and Edouard Grave. 2020.
\newblock \href {https://aclanthology.org/2020.lrec-1.494} {{CCN}et: Extracting high quality monolingual datasets from web crawl data}.
\newblock In \emph{Proceedings of the Twelfth Language Resources and Evaluation Conference}, pages 4003--4012, Marseille, France. European Language Resources Association.

\bibitem[{Wolf et~al.(2020)Wolf, Debut, Sanh, Chaumond, Delangue, Moi, Cistac, Rault, Louf, Funtowicz, Davison, Shleifer, von Platen, Ma, Jernite, Plu, Xu, Le~Scao, Gugger, Drame, Lhoest, and Rush}]{wolf-etal-2020-transformers}
Thomas Wolf, Lysandre Debut, Victor Sanh, Julien Chaumond, Clement Delangue, Anthony Moi, Pierric Cistac, Tim Rault, Remi Louf, Morgan Funtowicz, Joe Davison, Sam Shleifer, Patrick von Platen, Clara Ma, Yacine Jernite, Julien Plu, Canwen Xu, Teven Le~Scao, Sylvain Gugger, Mariama Drame, Quentin Lhoest, and Alexander Rush. 2020.
\newblock \href {https://doi.org/10.18653/v1/2020.emnlp-demos.6} {Transformers: State-of-the-art natural language processing}.
\newblock In \emph{Proceedings of the 2020 Conference on Empirical Methods in Natural Language Processing: System Demonstrations}, pages 38--45, Online. Association for Computational Linguistics.

\bibitem[{Xue et~al.(2020)Xue, Constant, Roberts, Kale, Al-Rfou, Siddhant, Barua, and Raffel}]{xue2020mt5}
Linting Xue, Noah Constant, Adam Roberts, Mihir Kale, Rami Al-Rfou, Aditya Siddhant, Aditya Barua, and Colin Raffel. 2020.
\newblock mt5: A massively multilingual pre-trained text-to-text transformer.
\newblock \emph{arXiv preprint arXiv:2010.11934}.

\bibitem[{Zhao et~al.(2011)Zhao, Jiang, Weng, He, Lim, Yan, and Li}]{zhao2011comparing}
Wayne~Xin Zhao, Jing Jiang, Jianshu Weng, Jing He, Ee-Peng Lim, Hongfei Yan, and Xiaoming Li. 2011.
\newblock Comparing twitter and traditional media using topic models.
\newblock In \emph{European conference on information retrieval}, pages 338--349. Springer.

\bibitem[{Zubiaga et~al.(2018)Zubiaga, Aker, Bontcheva, Liakata, and Procter}]{10.1145/3161603}
A.~Zubiaga, A.~Aker, K.~Bontcheva, M.~Liakata, and R.~Procter. 2018.
\newblock \href {https://doi.org/10.1145/3161603} {Detection and resolution of rumours in social media}.
\newblock \emph{ACM Computing Surveys}, 51:1--36.

\end{thebibliography}

\appendix

\section{Annotation Guidelines}
\label{sec:appendix-guidelines}
Below we provide the guidelines provided to the coders of each language.
\subsubsection{English}
Choose the appropriate topics expressed by the text. 
You can work on this task only once,  multiple tasks from the same annotators will be rejected. 
Some simple sentences are designed to verify the quality of  the annotations.. We will reject tasks where these simple test questions are not correct. 

For privacy reasons and to make the annotation easier, all non-verified user mentions are represented as \{\{USER\}\} and all URL entries as \{\{URL\}\}.

1. Arts \& Culture: Content about art forms, which evinces some degree of talent, training, or professionalism.

2. Business \& Entrepreneurs: Content that relates to money, the economy, and wealth creation broadly. Including job tips, career advice, and day in the life.

3. Celebrity \& Pop Culture: Stars and celebrities, their lives, funny moments, relationships, and fan communities.

4. Diaries \& Daily Life: Slice of life, everyday content that illustrates personal opinions, feelings, occasions, and lifestyles.

5. Family: Family dynamics, in-jokes, and everyday moments.

6. Fashion \& Style: Content about fashion, outfits, looks, shows, street style, collections, and designers. Both amateur and professional.

7. Film, TV \& Video: Traditional media and entertainment, including film, and tv, as well as content about Netflix and other streaming shows.

8. Fitness \& Health: Healthy living and the components thereof, including nutrition, exercise, progress, and wellness.

9. Food \& Dining: Anything related to food and food culture. Cooking, restaurants, food, reviews, technique, and ASMR. 

10. Learning \& Educational: Instructive, informative, educational content that teaches a fact, skill or topic.

11. News \& Social Concern: Awareness, activism, and discussion of societal issues and injustices contents that focus on coverage of newsworthy events, political and otherwise.

12. Relationships: Relationship dynamics, jokes, relatable moments, and the like between friend groups and romantic partners.

13. Science \& Technology: Content related to technology, natural phenomena, as well as knowledge and theories about the future and the universe.

14. Youth \& Student Life: Moments and memes of life at school and in the classroom, including teachers, events, and the like.

15. Music: Music performance, discussion, experiences and the like.

16. Gaming: Video games related content, gameplay, competition, culture and other games (e.g. board games). 

17. Sports: All depictions of sports (e.g. football, baseball, cricket, tennis, etc.). 

18. Travel \& Adventure: Vacations, travel tips, lodgings, means of conveyance, and the experience of travel.

19. Other Hobbies: Hobbies and personal interests not included in the topics above.

Multiple topics are allowed, please check ALL the relevant topics to the text, when the topic is mixed.
Make sure that you check at least one topic in each text.
 
Do you understand the instructions?

\subsubsection{Spanish}
Elija los temas apropiados expresados por el texto.
Sólo puede trabajar en esta tarea una vez, se rechazarán varias tareas de los mismos anotadores.
Algunas oraciones simples están diseñadas para verificar la calidad de las anotaciones. Rechazaremos las tareas en las que estas preguntas de prueba simples no sean correctas.

Por motivos de privacidad y para facilitar la anotación, todas las menciones de usuarios no verificados se representan como \{\{USUARIO\}\} y todas las entradas de URL como \{\{URL\}\}.

1. Arte y cultura: Contenido sobre formas de arte que demuestre algún grado de talento, capacitación o profesionalismo.

2. Negocios y emprendedores: Contenido relacionado con el dinero, la economía y la creación de riqueza en general. Incluyendo consejos de trabajo, de carrera u otros.

3. Celebridades y cultura pop: Estrellas y celebridades, sus vidas, momentos divertidos, relaciones y comunidades de admiradores.

4. Diarios y vida diaria: Contenido cotidiano y de vida diaria que ilustra opiniones personales, sentimientos, eventos y estilos de vida.

5. Familia: Dinámicas y referencias familiares, momentos cotidianos.

6. Moda y estilo: Contenido sobre moda, atuendos, looks, desfiles, estilo callejero, colecciones y diseñadores. Tanto amateur como profesional.

7. Cine, televisión y video: Medios tradicionales y de entretenimiento, incluidos cine y televisión, así como contenido sobre programas de streaming.

8. Estado físico y salud: Estilos de vida saludable y similar, incluida la nutrición, el ejercicio, el progreso y el bienestar.

9. Food \& Dining: Todo lo relacionado con la comida y la cultura gastronómica. Cocina, restaurantes, comida, reseñas, recetas y otros.

10. Aprendizaje y educación: Contenido instructivo, informativo y educativo para enseñar hechos, habilidades o temáticas.

11. Noticias e interés social: Conciencia, activismo y debate sobre problemas sociales y contenidos de injusticias que se centran en la cobertura de eventos de interés periodístico, políticos y de otro tipo.

12. Relaciones: Dinámicas de relación, bromas, momentos identificables y similares entre grupos de amigos y parejas románticas.

13. Ciencia y Tecnología: Contenido de tecnología, fenómenos naturales, así como conocimientos y teorías sobre el futuro y el universo.

14. Juventud y Vida Estudiantil: Momentos y memes de la vida en la escuela y en clase, incluidos maestros, eventos y similares.

15. Música: Interpretación musical, discusión, experiencias y similares.

16. Juegos: Contenido relacionado con videojuegos, juegos de rol, competición y otros juegos (por ejemplo, juegos de mesa).

17. Deportes: Todo lo relacionado con el deporte (por ejemplo, fútbol, béisbol, atletismo, tenis, etc.).

18. Viajes y aventuras: Vacaciones, consejos de viaje, alojamiento, medios de transporte y experiencias de viaje.

19. Otros pasatiempos: Pasatiempos, hobbies e intereses personales no incluidos en los temas anteriores.

Se permiten múltiples temas, marque TODOS los temas relevantes para el texto (puede ser más de uno cuando la temática es variada).

Asegúrese de marcar al menos un tema en cada texto.
 
¿Entiendes las instrucciones?

\subsubsection{Japanese}
\begin{CJK*}{UTF8}{min}
インストラクション

ツイートの文章に対し、適切なトピックをリストから選んでください。このアノテーションには一度しか参加することはできません。同じアノテーターから複数のアノテーションがあった場合、それは受理されることはありませんので注意してください。アノテーションの品質保持のためアノテーションの中にはいくつか簡単な例題があり、それらを間違えた場合もアノテーションは受理されません。

ツイートのプライバシー保護のため、non-verified user name 及び web url はマスキングされています。

1. アート\&カルチャー: アートや文化など芸術性や専門性の高い物に関するツイート。

2. ビジネス: 経済やビジネス、金融などに関わるツイート。キャリア形成や転職情報なども含まれます。

3. 芸能: 芸能人やそれらが主催するイベントなどに関するツイート。

4. 日常: 日々の出来事などの日常的な事柄に関するツイート。

5. 家族: 家族に関するツイート

6. ファッション: ストリートスナップやデザイン、ファッションに関するツイート。

7. 映画\&ラジオ: TVやラジオ、映画などのエンタメ等に関するツイート。

8. フィットネス\&健康: 栄養、フィットネスなどに関するツイート。

9. 料理: 料理やレストランなど食に関するツイート

10. 教育関連: 教育に関するツイート。

11. 社会: 社会情勢やそれに通ずるニュース、政治などに関するツイート。

12. 人間関係: パートナーシップや恋人との関係性などに関するツイート。

13. サイエンス: IT含むサイエンスに関するツイート。

14. 学校: 学校での出来事や行事に関するツイート。

15. 音楽: 音楽フェスや音楽そのものに関するツイート。

16. ゲーム: ゲーム（オンラインゲームやビデオゲーム等）に関するツイート。

17. スポーツ: スポーツに関するツイート。

18. 旅行: 旅行に関するツイート。

19. その他: その他、趣味や個人の嗜好に関するツイート。
一つのツイートに対し複数のラベルの付与が可能になってます。

少なくとも一つのトピックを選んでください。
 
インストラクションは理解できましたでしょうか？
\end{CJK*}

\subsubsection{Greek}
\selectlanguage{greek}
Επιλέξτε τα κατάλληλα θέματα που εκφράζει το κείμενο.

Μπορείτε να εργαστείτε σε αυτήν την εργασία μόνο μία φορά, πολλές εργασίες από τους ίδιους σχολιαστές θα απορριφθούν. Ορισμένες απλές προτάσεις έχουν σχεδιαστεί για να επαληθεύουν την ποιότητα των σχολιασμών. Θα απορρίψουμε εργασίες όπου αυτές οι απλές ερωτήσεις δοκιμής δεν είναι σωστές.
Για λόγους απορρήτου και για να γίνει ευκολότερος ο σχολιασμός, όλες οι μη επαληθευμένες αναφορές χρηστών αντιπροσωπεύονται ως \selectlanguage{english} \{\{USER\}\} \selectlanguage{greek} και όλες οι \selectlanguage{english} URL \selectlanguage{greek} ως \selectlanguage{english} \{\{URL\}\}.

\selectlanguage{greek}
1. Τέχνες \& Πολιτισμός: Περιεχόμενο για μορφές τέχνης, το οποίο δείχνει κάποιο βαθμό ταλέντου, κατάρτισης ή επαγγελματισμού.

2. Επιχειρήσεις \& Επιχειρηματίες: Περιεχόμενο που σχετίζεται γενικά με τα χρήματα, την οικονομία και τη δημιουργία πλούτου. Συμπεριλαμβάνονται συμβουλές για δουλειά, συμβουλές σταδιοδρομίας, κτλ.

3. Διασημότητες \& Ποπ κουλτούρα: Αστέρια και διασημότητες, η ζωή τους, αστείες στιγμές, σχέσεις και κοινότητες θαυμαστών.

4. Ημερολόγια \& Καθημερινή ζωή: Στιγμές της ζωής, καθημερινό περιεχόμενο που απεικονίζει προσωπικές απόψεις, συναισθήματα, περιστάσεις και τρόπους ζωής.

5. Οικογένεια: Δυναμική της οικογένειας, αστεία και καθημερινές στιγμές.

6. Μόδα \& Στυλ: Περιεχόμενο σχετικά με τη μόδα, τα ρούχα, τις εμφανίσεις, τις επιδείξεις, το street style, τις συλλογές και τους σχεδιαστές. Ερασιτεχνική και επαγγελματική.

7. Ταινίες, τηλεόραση \& βίντεο: Παραδοσιακά μέσα και ψυχαγωγία, συμπεριλαμβανομένων ταινιών και τηλεόρασης, καθώς και περιεχόμενο για το Netflix και άλλες εκπομπές ροής.

8. Γυμναστική \& Υγεία: Υγιεινή ζωή και τα συστατικά της, συμπεριλαμβανομένης της διατροφής, της άσκησης, της προόδου και της ευεξίας.

9. Φαγητό \& Δείπνο: Οτιδήποτε σχετίζεται με το φαγητό και την κουλτούρα του φαγητού. Μαγειρική, εστιατόρια, φαγητό, κριτικές, τεχνική και \selectlanguage{english}ASMR\selectlanguage{greek}.

10. Μάθηση \& Εκπαίδευση: Εκπαιδευτικό, ενημερωτικό, εκπαιδευτικό περιεχόμενο που διδάσκει ένα γεγονός, μια δεξιότητα ή ένα θέμα.

11. Ειδήσεις \& Κοινωνία: Ευαισθητοποίηση, ακτιβισμός και συζήτηση για κοινωνικά ζητήματα και αδικίες, περιεχόμενα που εστιάζουν στην κάλυψη γεγονότων άξιων ειδήσεων, πολιτικών και άλλων.

12. Σχέσεις: Δυναμική σχέσεων, αστεία, συγγενείς στιγμές και άλλα παρόμοια μεταξύ ομάδων φίλων και ρομαντικών συντρόφων.

13. Επιστήμη \& Τεχνολογία: Περιεχόμενο αιχμής τεχνολογίας, φυσικά φαινόμενα, καθώς και γνώση και θεωρίες για το μέλλον και το σύμπαν.

14. Νεανική \& Φοιτητική ζωή: Στιγμές και memes της ζωής στο σχολείο και στην τάξη, συμπεριλαμβανομένων δασκάλων, εκδηλώσεων και παρόμοια.

15. Μουσική: Μουσική παράσταση, συζήτηση, εμπειρίες και παρόμοια.

16. Παιχνίδια: περιεχόμενο σχετικό με βιντεοπαιχνίδια, παιχνίδι, ανταγωνισμό, πολιτισμό και άλλα παιχνίδια (π.χ. επιτραπέζια παιχνίδια).

17. Αθλητισμός: Όλες οι απεικονίσεις αθλημάτων (π.χ. ποδόσφαιρο, μπέιζμπολ, τένις).

18. Ταξίδια \& Περιπέτεια: Διακοπές, ταξιδιωτικές συμβουλές, καταλύματα, μεταφορικά μέσα και η εμπειρία του ταξιδιού.

19. Άλλα χόμπι: Χόμπι και προσωπικά ενδιαφέροντα που δεν περιλαμβάνονται στα παραπάνω θέματα.

Επιτρέπονται πολλά θέματα, παρακαλούμε ελέγξτε ΟΛΑ τα σχετικά θέματα στο κείμενο, όταν τα θέματα αναμιγνύονται.
Βεβαιωθείτε ότι έχετε επιλέξει τουλάχιστον ένα θέμα σε κάθε κείμενο.

Καταλαβαίνετε τις οδηγίες;

\selectlanguage{english}

\section{Models \& Dataset}
\label{sec:appendix-models}

\subsection{Dataset}
Table \ref{tab:preprocessing_steps} displays the number of remaining tweets in each preprocessing step for each language. The steps are: 1) language detection (ftext), 2) removal of incomplete/abusing tweets, 3) deduplication, 4) removal of tweets with high ammount of mentions and emojis, and 5) removal of tweets containing URLs.

\begin{table*}[]
\begin{tabular}{c|cccccc}
   & Total   & ftext   & incomplete/abusing & deduplication & mentions/emojis & URLS    \\ \hline
en & 225,400 & 217,491 & 208,442            & 193,560       & 178,841         & 81,929  \\ \hline
es & 225,350 & 218,163 & 197,617            & 186,266       & 178,060         & 110,669 \\ \hline
ja & 455,846 & 455,846 & 438,080            & 407,589       & 383,669         & 207,472 \\ \hline
gr & 225,300 & 218,461 & 214,031            & 206,147       & 203,947         & 30,858 
\end{tabular}
\caption{Number of remaining tweets for each preprocessing step for every language.}
\label{tab:preprocessing_steps}
\end{table*}

\begin{figure}
    \centering
    \includegraphics[width=1\linewidth]{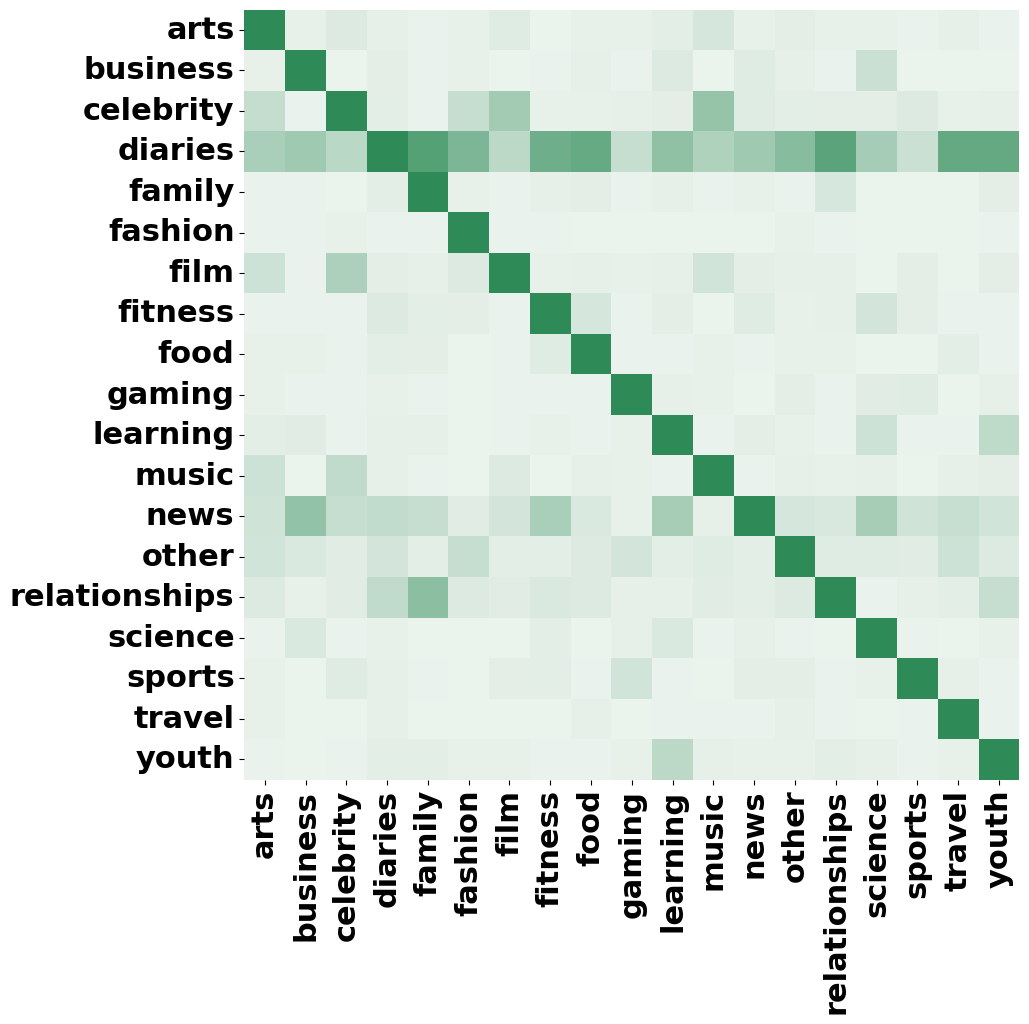}
    \caption{Overlap between topics across all languages. Darker color indicates higher overlap}
    \label{fig:overlap}
\end{figure}

Figure \ref{fig:overlap} displays the overlap between topics across all languages.
\label{sec:appendix-dataset}

\subsection{Models}
In total we estimate 168 hours used for the training of  \textit{bernice, xlm\_r}, and \textit{xlm\_t} models using a NVIDIA GeForce RTX 4090  GPU and 20 hours for \textit{bloomz} and \textit{mt0} models using an  NVIDIA Quadro RTX 8000 GPU.
Table \ref{tab:parameters} provides details for the models used in our experiments.

\begin{table}[ht]
\centering
\begin{tabular}{|l|l|}
\hline
\textbf{Model}         & \textbf{Parameters} \\ \hline
Bernice                              & 125M                  \\ \hline
XLM-R(T) base                           & 270M                  \\ \hline
XLM-R(T) large                           & 550M  \\ \hline
bloomz & 7B \\ \hline
mt0                           & 13B  \\ \hline
chat-gpt                      &  175B (approximate)     \\ \hline
\end{tabular}
\caption{Number of Parameters in different language models used.}
\label{tab:parameters}
\end{table}

\subsection{Prompts}
\label{sec:appendix-prompts}
Below we present the prompt used in the zero and few-shot settings of our experiments. The prompt used were similar to the ones used in \citet{muennighoff2022crosslingual}.

\noindent Classify the text "\{\{ tweet \}\}" into the following topics:
- \{\{ answer\_choices | join('\textbackslash n- ') \}\}

\noindent Topics:

\subsection{Topics Abbreviation}
\label{sec:appendix-topic_abbreviations}
Below we provide the abbreviations of topics used in the paper:

Arts \& Culture: arts

Business \& Entrepreneurs: business

Celebrity \& Pop Culture: celebrity

Diaries \& Daily Life: diaries

Family: family

Fashion \& Style: fashion

Film, TV \& Video: film

Fitness \& Health: fitness

Food \& Dining: food

Learning \& Educational: learning

News \& Social Concern: news

Relationships: relationships

Science \& Technology: science

Youth \& Student Life: youth

Music: music

Gaming: gaming

Sports: sports

Travel \& Adventure: travel

Other Hobbies: other

\section{Extended Results}
\label{sec:appendix-results}

\begin{figure*}
    \centering
    \scalebox{0.9}{
    \centering
    \includegraphics[width=\linewidth]{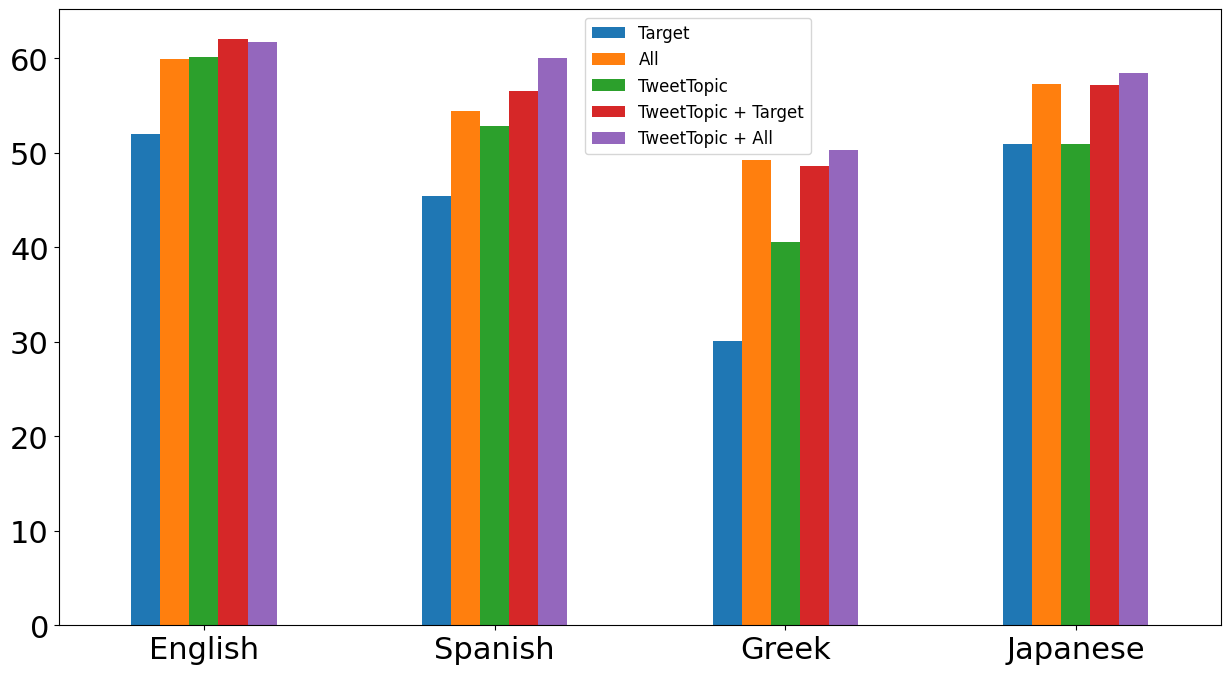}}
    \caption{F1 scores (macro average) of the best overall performing model (\textit{xlmt\_large}) in each setting and language.}
    \label{fig:best_f1}
\end{figure*}

\dimos{Figure \ref{fig:best_f1}, displays the scores achieved by the overall best-performing model, \textit{xlm\_t-large}, in each language and setting.}

\dimos{Tables \ref{tab:xlmt-details} and \ref{tab:gpt-4o-details} display detail results for the two best performing models, \textit{xlmt\_large} , trained on \textit{TweetTopic} and \textit{All languages}, and  \textit{gpt-4o}, in the \textit{few-shot} setting, respectively. The precision, recall, and f1 scores for each topic in every language are displayed. }

\begin{table}[h!]
    \centering
    \begin{tabular}{lcccc}
        \hline
        \textbf{Metric} & \textbf{en} & \textbf{es} & \textbf{ja} & \textbf{gr} \\
        \hline
        macro & 6.4 & 6.5 & 1.5 & 5.0 \\
        micro & 30.4 & 44.0 & 8.3 & 7.4 \\
        \hline
    \end{tabular}
    \caption{Macro and F1 scores for each language for the SuperCTM model.}
    \label{tab:ctopic}
\end{table}

\begin{table*}[]
\begin{tabular}{l|ccc|ccc|ccc|ccc}
\textbf{} &
  \multicolumn{3}{c|}{\textbf{en}} &
  \multicolumn{3}{c|}{\textbf{es}} &
  \multicolumn{3}{c|}{\textbf{gr}} &
  \multicolumn{3}{c}{\textbf{ja}} \\ \hline
\textit{topic} &
  \textbf{Pr} &
  \textbf{Rec} &
  \textbf{F1} &
  \textbf{Pr} &
  \textbf{Rec} &
  \textbf{F1} &
  \textbf{Pr} &
  \textbf{Rec} &
  \textbf{F1} &
  \textbf{Pr} &
  \textbf{Rec} &
  \textbf{F1} \\ \hline
arts \& culture           & 26 & 20 & 23 & 60 & 34 & 40 & 48 & 42 & 44 & 32 & 19 & 24 \\
business \& entrepreneurs & 79 & 65 & 70 & 55 & 34 & 41 & 51 & 36 & 41 & 64 & 45 & 52 \\
celebrity \& pop culture  & 54 & 49 & 51 & 60 & 57 & 57 & 48 & 42 & 43 & 60 & 70 & 64 \\
diaries \& daily life     & 80 & 71 & 75 & 77 & 85 & 81 & 70 & 81 & 75 & 80 & 83 & 81 \\
family                    & 85 & 60 & 69 & 60 & 58 & 59 & 66 & 78 & 71 & 57 & 50 & 53 \\
fashion \& style          & 70 & 70 & 69 & 80 & 65 & 68 & 56 & 50 & 52 & 40 & 30 & 33 \\
film tv \& video          & 73 & 74 & 73 & 46 & 51 & 47 & 61 & 65 & 62 & 67 & 66 & 66 \\
fitness \& health         & 69 & 54 & 57 & 74 & 52 & 60 & 79 & 65 & 72 & 62 & 62 & 62 \\
food \& dining            & 91 & 72 & 79 & 95 & 78 & 83 & 87 & 87 & 87 & 68 & 44 & 51 \\
gaming                    & 82 & 61 & 67 & 50 & 60 & 53 & 66 & 68 & 66 & 13 & 10 & 11 \\
learning \& educational   & 59 & 22 & 30 & 52 & 55 & 52 & 60 & 63 & 52 & 70 & 58 & 62 \\
music                     & 79 & 87 & 82 & 73 & 80 & 76 & 69 & 72 & 69 & 75 & 53 & 58 \\
news \& social concern    & 76 & 68 & 72 & 88 & 90 & 89 & 51 & 33 & 40 & 91 & 89 & 90 \\
other hobbies             & 43 & 26 & 32 & 37 & 17 & 23 & 43 & 43 & 43 & 23 & 13 & 14 \\
relationships             & 82 & 62 & 71 & 78 & 73 & 75 & 54 & 46 & 50 & 63 & 57 & 60 \\
science \& technology     & 66 & 68 & 67 & 90 & 65 & 71 & 38 & 33 & 34 & 63 & 33 & 39 \\
sports                    & 87 & 93 & 90 & 84 & 79 & 81 & 81 & 73 & 75 & 95 & 92 & 93 \\
travel \& adventure       & 68 & 50 & 57 & 63 & 32 & 39 & 65 & 56 & 60 & 27 & 29 & 25 \\
youth \& student life     & 46 & 31 & 37 & 64 & 37 & 44 & 87 & 76 & 78 & 31 & 12 & 17
\end{tabular}
\caption{Precision (Pr), Recall (Rec), and F1 scores for each topic achieved by  \textit{xlmt\_large} trained on  \textit{TweetTopic} and \textit{All languages}.}
\label{tab:xlmt-details}
\end{table*}

\begin{table*}[]
\begin{tabular}{l|ccc|ccc|ccc|ccc}
 & \multicolumn{3}{c|}{en} & \multicolumn{3}{c|}{es} & \multicolumn{3}{c|}{gr} & \multicolumn{3}{c}{ja} \\ \hline
\textit{topic} & Pr & Rec & F1 & Pr & Rec & F1 & Pr & Rec & F1 & Pr & Rec & F1 \\ \hline
arts \& culture & 52 & 28 & 36 & 65 & 34 & 39 & 55 & 40 & 44 & 61 & 28 & 38 \\
business \& entrepreneurs & 72 & 50 & 58 & 79 & 27 & 39 & 88 & 34 & 44 & 47 & 13 & 20 \\
celebrity \& pop culture & 50 & 65 & 56 & 50 & 58 & 53 & 70 & 55 & 61 & 51 & 47 & 46 \\
diaries \& daily life & 86 & 40 & 55 & 91 & 38 & 54 & 93 & 50 & 65 & 76 & 60 & 67 \\
family & 82 & 67 & 73 & 49 & 63 & 52 & 45 & 56 & 50 & 58 & 62 & 59 \\
fashion \& style & 55 & 93 & 68 & 39 & 70 & 46 & 47 & 50 & 47 & 41 & 55 & 46 \\
film tv \& video & 86 & 69 & 76 & 57 & 39 & 46 & 95 & 49 & 64 & 63 & 63 & 62 \\
fitness \& health & 57 & 65 & 60 & 62 & 37 & 45 & 58 & 47 & 49 & 82 & 53 & 64 \\
food \& dining & 95 & 62 & 73 & 75 & 79 & 76 & 67 & 69 & 66 & 80 & 73 & 76 \\
gaming & 60 & 69 & 63 & 40 & 48 & 42 & 40 & 30 & 33 & 76 & 70 & 73 \\
learning \& educational & 55 & 27 & 37 & 83 & 88 & 85 & 63 & 33 & 42 & 61 & 62 & 55 \\
music & 73 & 88 & 80 & 82 & 50 & 62 & 69 & 77 & 68 & 61 & 74 & 66 \\
news \& social concern & 71 & 71 & 71 & 86 & 59 & 67 & 95 & 86 & 90 & 60 & 29 & 38 \\
other hobbies & 51 & 16 & 24 & 40 & 29 & 31 & 20 & 4 & 7 & 48 & 31 & 38 \\
relationships & 83 & 59 & 69 & 66 & 89 & 75 & 69 & 48 & 57 & 68 & 26 & 37 \\
science \& technology & 69 & 62 & 65 & 30 & 60 & 40 & 60 & 29 & 36 & 20 & 35 & 25 \\
sports & 88 & 96 & 92 & 73 & 88 & 80 & 93 & 95 & 94 & 79 & 85 & 82 \\
travel \& adventure & 59 & 52 & 54 & 50 & 42 & 45 & 37 & 42 & 35 & 66 & 67 & 63 \\
youth \& student life & 47 & 34 & 39 & \multicolumn{1}{l}{22} & \multicolumn{1}{l}{15} & \multicolumn{1}{l|}{17} & 35 & 12 & 18 & 52 & 49 & 49
\end{tabular}
\caption{Precision (Pr), Recall (Rec), and F1 scores for each topic achieved by  \textit{gpt-4o} in the few-shot setting.}
\label{tab:gpt-4o-details}
\end{table*}

\dimos{Table \ref{tab:ctopic} displays the macro and micro F1 scores achieved when using supervised SuperCTM  \cite{card2017neural} with the default parameters as provided in the Contextualized Topic Models (CTM) \cite{bianchi-etal-2021-cross} implementation. The model was trained using both TweetTopic and {\DATASET}. As seen by the results the model fails to perform well and only manages to achieve mediocre micro-F1 scores when tested on English and Spanish.}

\end{document}